\documentclass{WileyMSP-template}
\usepackage{lipsum}
\usepackage[table,xcdraw]{xcolor} 
\usepackage{graphicx}
\usepackage{amsmath,amssymb,amsfonts,bm}
\usepackage[style=ieee, url=false]{biblatex}
\usepackage[normalem]{ulem}
\usepackage{booktabs}
\usepackage{multirow}
\usepackage{multicol}

\DeclareSourcemap{
  \maps{
    \map{
      \pertype{article}
      \step[fieldset=language, null]
      \step[fieldset=url, null]
      \step[fieldset=doi, null]
      \step[fieldset=issn, null]
      \step[fieldset=isbn, null]
      \step[fieldset=note, null]
      \step[fieldset=editor, null]
      \step[fieldset=urldate, null]
      \step[fieldset=file, null]
    }
  }
}
\DeclareSourcemap{
  \maps{
    \map{
      \pertype{inproceedings}
      \step[fieldset=language, null]
      \step[fieldset=url, null]
      \step[fieldset=doi, null]
      \step[fieldset=issn, null]
      \step[fieldset=isbn, null]
      \step[fieldset=note, null]
      \step[fieldset=editor, null]
      \step[fieldset=urldate, null]
      \step[fieldset=file, null]
    }
  }
}
\DeclareSourcemap{
  \maps{
    \map{
      \pertype{incollection}
      \step[fieldset=language, null]
      \step[fieldset=url, null]
      \step[fieldset=doi, null]
      \step[fieldset=issn, null]
      \step[fieldset=isbn, null]
      \step[fieldset=note, null]
      \step[fieldset=editor, null]
      \step[fieldset=urldate, null]
      \step[fieldset=file, null]
    }
  }
}
\DeclareSourcemap{
  \maps{
    \map{
      \pertype{misc}
      \step[fieldset=language, null]
      \step[fieldset=url, null]
      \step[fieldset=doi, null]
      \step[fieldset=issn, null]
      \step[fieldset=isbn, null]
      \step[fieldset=note, null]
      \step[fieldset=editor, null]
      \step[fieldset=urldate, null]
      \step[fieldset=file, null]
    }
  }
}

\newif\ifshowmodifications
\newif\ifshowmodificationsVtwo
\showmodificationsfalse
\showmodificationsVtwofalse

\newcommand{\modify}[1]{%
  \ifshowmodifications
    {\color{blue}#1}%
  \else
    {\color{black}#1}%
  \fi
}

\newcommand{\modifyVtwo}[1]{%
  \ifshowmodificationsVtwo
    {\color{blue}#1}%
  \else
    {\color{black}#1}%
  \fi
}

\begin{document}

\pagestyle{fancy}
\rhead{\includegraphics[width=2.5cm]{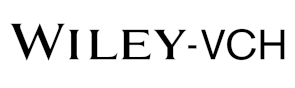}}

\title{Crater Observing Bio-inspired Rolling Articulator (COBRA)}
\maketitle

\author{Adarsh Salagame}
\author{Henry Noyes}
\author{Eric Sihite}
\author{Arash Kalantari}
\author{Alireza Ramezani*}


\begin{affiliations}
\medskip
Adarsh Salagame, Henry Noyes, Prof. Alireza Ramezani\\
Northeastern University, Boston, MA, USA\\
Email Address: salagame.a, noyes.he, a.ramezani@northeastern.edu \\
\medskip
Dr. Eric Sihite\\
California Institute of Technology, Pasadena, CA, USA\\
Email Address: esihite@caltech.edu\\
\medskip
Dr. Arash Kalantari\\
Jet Propulsion Laboratory, Pasadena, CA, USA\\
Email Address: arash.kalantari@jpl.nasa.gov
\end{affiliations}


\keywords{Bio-inspired, Mobile Robot, Space Exploration}

\begin{abstract}

NASA aims to establish a sustainable human basecamp on the Moon as a stepping stone for future missions to Mars and beyond. The discovery of water ice on the Moon's craters located in permanently shadowed regions, which can provide drinking water, oxygen, and rocket fuel, is therefore of critical importance. However, current methods to access lunar ice deposits are limited. While rovers have been used to explore the lunar surface for decades, they face significant challenges in navigating harsh terrains, such as permanently shadowed craters, due to the high risk of immobilization. This report introduces COBRA (Crater Observing Bio-inspired Rolling Articulator), a multi-modal snake-style robot designed to overcome mobility challenges in Shackleton Crater's rugged environment. COBRA combines slithering and tumbling locomotion to adapt to various crater terrains. In snake mode, it uses sidewinding to traverse flat or low inclined surfaces, while in tumbling mode, it forms a circular barrel by linking its head and tail, enabling rapid movement with minimal energy on steep slopes. Equipped with an onboard computer, stereo camera, inertial measurement unit, and joint encoders, COBRA facilitates real-time data collection and autonomous operation. This paper highlights COBRA’s robustness and efficiency in navigating extreme terrains through both simulations and experimental validation.


\end{abstract}




\section{Introduction}

Robots have become essential tools in our scientific exploration of space and celestial bodies. Equipped with advanced sensors and designed to withstand severe conditions, they safely explore, sample, and analyze, yielding invaluable insights that would otherwise remain unattainable. Traditionally, robots for space exploration have relied on wheeled locomotion \cite{thoesen_planetary_2021}. However, recent innovations have shifted this paradigm. Ingenuity, the Mars Helicopter \cite{balaram_ingenuity_2021}, has successfully demonstrated the first aerial locomotion on another planet, while alternate surface mobility systems such as Dragonfly, EELS, and DuAxel \cite{turtle_dragonfly_2024, vaquero_eels_2024, mcgarey_towards_2019} are in advanced stages of development. These solutions are typically tailored to address the locomotion challenges in specific regions of interest, such as recurring slope lineae on Mars for DuAxel or the icy moon Enceladus for EELS. 

However, some regions of interest, particularly the Moon's permanently shadowed regions (PSRs), are still unexplored due to limitations of existing surface mobility solutions. In this work, we introduce our bio-inspired mobile robot, COBRA, Crater Observing Bio-inspired Rolling Articulator, designed to address challenges unmet by current solutions for energy-efficient exploration of steep and unknown craters in the Moon's PSRs. As a case study, we focus on Shackleton Crater (see Fig.~\ref{fig:mission-scenario}) at the lunar South Pole, a site of immense scientific potential for water ice discovery \cite{zuber_constraints_2012}, but significant challenges for existing exploration methods. We present the envisioned mission scenario and the locomotion solution that led to COBRA winning the 2022 Breakthrough, Innovative, and Game-changing (BIG) Idea Challenge~\cite{bigidea}, an initiative under NASA’s Space Technology Mission Directorate’s (STMD’s) Game Changing Development Program (GCD), for alternative lunar mobility solutions. This report summarizes the design, dynamical modeling, contact-rich and optimization-based gait discovery, as well as the numerical and experimental evaluation of COBRA's mobility in unstructured, contested environments. Specifically, the technical contributions of this work are: \modify{ (1) Hardware design, including head, tail, and body module mechanical and electronics design. (2) Design of the head-tail docking mechanism to substantiate transitions between crawling and tumbling. (3) Numerical modeling of crawling and tumbling locomotion feats based on Lagrange and mixed Hamilton-Lagrange (partitioned state-space model for tumbling) dynamics. (4) Contact-implicit optimization-based gait discovery and joint motion planning. (5) Simulation and experimental validation of hex-ring tumbling, spiral tumbling, sidewinding, lateral rolling, and vertical undulation. (6) Demonstration of field-tested locomotion modes in dusty, steep, and bumpy environments, with self-sustained dust mitigation. (7) Demonstration of full deployment from a lander, including unmanned transitions between various locomotion modes in experiments.}


\subsection{Importance of Shackleton Crater}
With the Artemis program, NASA and collaborating space agencies aim to revitalize lunar and space exploration. One Artemis objective is to create a sustainable human base camp on the Moon with the hope to then propel further missions to Mars and beyond. Accomplishing such a feat is contingent on In-Situ Resource Utilization (ISRU) on the Moon \cite{sanders_progress_2012}. One resource of significant interest is lunar water ice, which can potentially supply drinking water, oxygen, and rocket propellant. However, the means to access the ice deposits on the Moon still require additional development \cite{reid_artemis_2015,just_parametric_2020,zacny_pneumatic_2008}. In 2018, NASA confirmed the presence of water ice on the Moon’s poles, concentrated chiefly in PSRs \cite{liu_integrated_2020}. The near-permanent lack of sunlight in these regions results in extremely low temperatures (as low as -238 °C) and allows for the accumulation of water ice and other volatiles \cite{slyuta_physical_2014}. 
One such PSR is Shackleton Crater.

While the presence of water ice in Shackleton Crater is evident \cite{zuber_constraints_2012}, there are no precise measurements of its quantity or chemical composition. This detailed information, along with topographic maps of crater terrain, is critical to initiate targeted mining operations for ISRU. However, acquiring these measurements requires proximate investigations in extreme lunar environments that pose significant mobility challenges to exploration platforms.

\subsubsection*{Locomotion Challenges} 

\textit{High Porosity Regolith Surface:} The surface of the lunar South Pole is characterized by high porosity regolith that poses significant locomotion challenges \cite{suescun-florez_geotechnical_2015}. Due to the high porosity, traditional wheeled rovers suffer sinkage and slipping, reducing energy efficiency and increasing the risk of immobilization. The high porosity lunar dust is also abrasive and invades machinery due to its particulate nature \cite{connors_interviews_1994}.

\textit{Terrain Profile:} Traversing the immense slopes to reach all areas of scientific interest inside the crater also poses significant challenges. Shackleton Crater is a massive geographic feature 21 kilometers in diameter. The crater slope leading to the crater floor has an average slope of 30.5 degrees and covers a horizontal distance of approximately 8 kilometers. This steep crater slope is difficult to both ascend and descend. On the descent, there is limited traction due to the lower lunar gravity further reducing normal force, and the regolith substrate being more prone to yield, leading to slipping and sinkage. 

This mobility challenge also extends to the uneven crater floor and surrounding areas. The crater slope and floor have a root mean square (RMS) surface roughness of approximately 1 meter. This makes it difficult for wheeled systems to traverse, as the height of the obstacle they can overcome is limited by their wheel diameter. Boulder fields outside the crater, the unknown mechanical properties of regolith with ice, and a lack of detailed topography of the crater floor further raise the requirements for an adaptable system.

\modify{ A large of number of studies have investigated the mechanics of wheeled and legged locomotion in such conditions, using both analytical terramechanics models and experimental proxies. For instance, Ishigami et al. \cite{ishigami_terramechanics-based_2007} and Shrivastava et al. \cite{shrivastava_material_2020} analyze wheel-soil interactions using lunar simulants, including deformable media like poppy seeds, to understand sinkage and traction loss. Similarly, recent work by Kolvenbach et al. \cite{kolvenbach_traversing_2021} and Karsai et al. \cite{karsai_real-time_2022} explores how legged robots navigate granular slopes, addressing challenges in slip, footing stability, and dynamic control. However, the systematic consideration of substrate phenomena--such as mass wasting, sinkage, and traction loss--during inclined traversal remains largely unexplored.}

\textit{Lack of sunlight:} Shackleton Crater is a PSR, leading to two challenges: power generation and near-absolute zero temperature. Traversal of the crater is limited due to the lack of solar power. Therefore, large distances must be covered with minimal power consumption, or systems must rely on heavy power generation mechanisms such as radioisotope thermoelectric generators (RTGs). Low temperatures interfere with mechanisms that rely on lubricants and liquid batteries. This results in less efficient power systems. 

\subsection{State Of The Art and Research Gap}
To overcome locomotion challenges in the Moon's hostile environment, a robotic system must be designed to minimize sinkage and slippage, while incorporating features to prevent immobilization. It must traverse long distances in an energy-efficient manner while maintaining passive and active regolith mitigation strategies and be able recover from, or continue to operate with, component failures.

\begin{figure}
    \centering
    \includegraphics[width=0.95\linewidth]{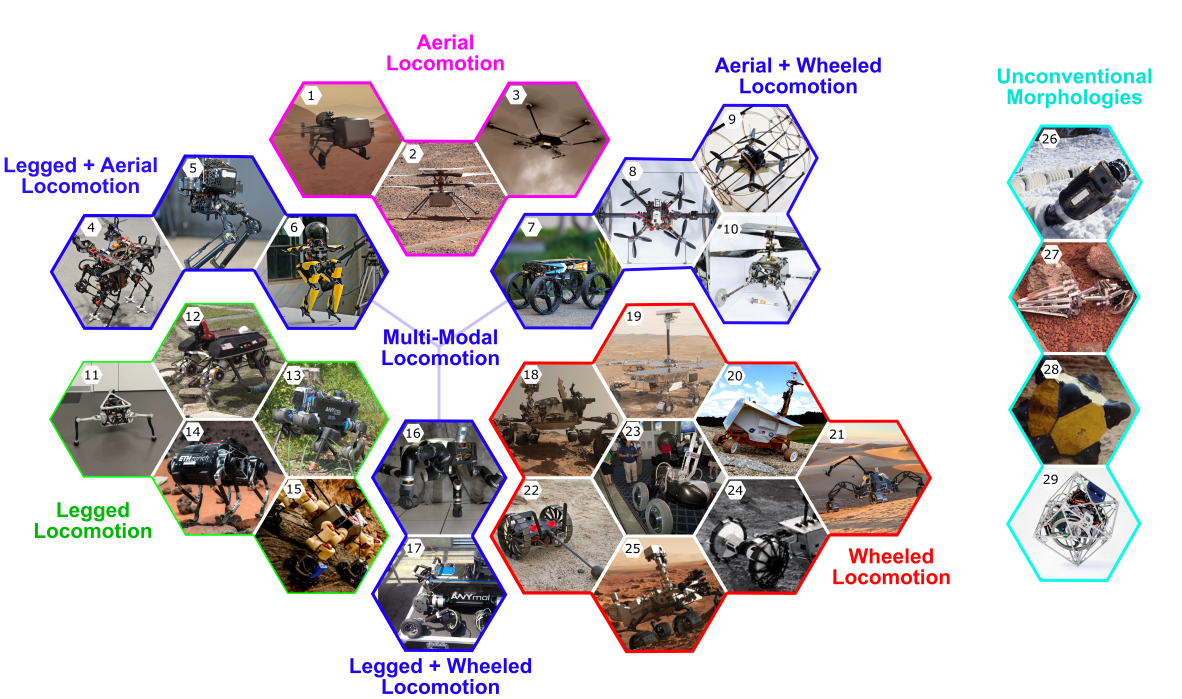}
    \caption{The composite figure shows examples of mobility systems for planetary surface exploration using various morphologies. (In Order: (1) Dragonfly \cite{turtle_dragonfly_2024}, (2) Ingenuity Mars Helicopter \cite{tzanetos_ingenuity_2022}, (3) Mars Science Helicopter \cite{tzanetos_future_2022}, (4) Husky Carbon \cite{salagame_letter_2022}, (5) Harpy \cite{dangol_thruster-assisted_2020}, (6) Leonardo \cite{kim2021bipedal}, (7) M4 \cite{sihite_multi-modal_2023}, (8) Rollocopter \cite{noauthor_rollocopter_nodate}, (9) HyTAQ \cite{kalantari_design_2013}, (10) Sample Recovery Helicopter \cite{mier-hicks_sample_2023}, (11) SpaceHopper \cite{spiridonov_spacehopper_2024}, (12) Llama \cite{nicholson_llama_2020}, (13) ANYmal \cite{hutter_anymal_2016}, (14) SpaceBok \cite{arm_spacebok_2019}, (15) LEMUR 3 \cite{parness_lemur_2017}, (16) Robosimian \cite{noauthor_jpl_nodate}, (17) Wheeled-Anymal \cite{medeiros_trajectory_2020}, (18) Curiosity Rover \cite{noauthor_mars_curiosity}, (19) Rosalind Franklin ExoMars Rover \cite{noauthor_exomars_nodate}, (20) VIPER \cite{noauthor_viper_nodate}, (21) SherpaTT \cite{cordes_sherpatt_2016}, (22) PUFFER \cite{bell_dynamic_2021}, (23) Scarab \cite{bartlett_design_2008}, (24) DuAxel \cite{matthews_design_2012}, (25) Perseverence Rover \cite{noauthor_mars_perseverance}, (26) EELS \cite{vaquero_eels_2024}, (27) Frogbot \cite{noauthor_frogbot_nodate}, (28) Hedgehog \cite{httpswwwjplnasagov_hedgehog_nodate}, (29) Cubli \cite{gajamohan_cubli_2012}).}
    \label{fig:lit-review}
\end{figure}

Current state-of-the-art wheeled rovers \cite{noauthor_mars_curiosity,noauthor_mars_perseverance, noauthor_viper_nodate, noauthor_exomars_nodate} shown in Fig.~\ref{fig:lit-review} struggle on steep inclines due to inadequate traction on the porous regolith and decreased stability from the reduced normal force. Hybrid systems such as SherpaTT \cite{cordes_sherpatt_2016} and Scarab \cite{bartlett_design_2008} partially address this issue by incorporating articulated legs that can lift the wheels off the ground to reposition them. More advanced multi-modal platforms like Robosimian \cite{noauthor_jpl_nodate} combine wheels with fully articulated legs, enabling both wheeled and legged locomotion, which improves steep slope traversal but remains slow and energy-intensive. An alternative approach is to use tether-assisted mobility, implemented in the DuAxel system \cite{nesnas_axel_2012}, where one half of the rover anchors itself while the other rappels down steep terrain using a tether. While effective in certain terrains, the scale and steepness of craters such as Shackleton limit the practicality of tethered solutions.

Legged systems, such as ETH Zurich’s Spacebok \cite{arm_spacebok_2019}, JPL's Llama \cite{nicholson_llama_2020}, and ANYmal \cite{hutter_anymal_2016}, offer another solution by allowing precise control of foot placement, enabling stable locomotion on loose surfaces. Despite their adaptability, these systems are generally energy-intensive. To address this, recent works like SpaceHopper \cite{spiridonov_spacehopper_2024} have explored hopping-based locomotion to leverage the Moon's low gravity for energy-efficient travel over long distances. For steeper slopes, climbing robots equipped with micro-grippers, such as JPL’s LEMUR 3 \cite{parness_lemur_2017}, have been developed, but these designs are tailored for hard, rocky surfaces and are unsuitable for the soft, granular lunar regolith.

Flying systems can traverse large distances quickly, as demonstrated by Ingenuity \cite{marsnasagov_mars_nodate} on Mars. NASA’s Dragonfly \cite{turtle_dragonfly_2024}, the Mars Science Helicopter \cite{tzanetos_future_2022} and the Sample Return Helicopter \cite{mier-hicks_sample_2023} are further exploring flight-based solutions for planetary exploration. However, replicating this approach on the Moon, which lacks an atmosphere, would require the use of ion or hydrogen-based propulsion systems. These technologies are extremely energy-intensive, making frequent recharging or carrying significant fuel reserves necessary for sustained operation. Thruster-assisted multi-modal designs like LEONARDO \cite{kim2021bipedal}, which combine hopping and flight, present a potential compromise, but their exhaust may alter the chemical composition of the lunar regolith, complicating scientific investigations. Similarly, other multi-modal systems such as the M4 \cite{sihite_multi-modal_2023} and JPL’s Rollocopter \cite{noauthor_rollocopter_nodate}, which combine wheeled mobility and flight, would encounter similar challenges.

Unconventional morphologies, such as the screw-driven snake robot EELS \cite{vaquero_eels_2024} designed for icy terrains, and micro-scale jumping robots like Frogbot \cite{noauthor_frogbot_nodate}, offer new paradigms for locomotion. To achieve safe locomotion down steep crater walls, a particularly promising approach is seen in JPL’s Hedgehog \cite{httpswwwjplnasagov_hedgehog_nodate}, a tumbling robot that uses internal flywheels to generate momentum. Tumbling locomotion takes advantage of gravity to descend slopes, eliminating the need for active actuation and resulting in highly energy-efficient traversal. Despite its long history, tumbling has seen limited application in space exploration, leaving its full potential largely unexplored. Given its simplicity and efficiency, it presents an exciting opportunity for navigating the Moon’s challenging terrain.

The merits and limitations of tumbling robots are well documented. Passive systems like NASA/JPL’s Mars Tumbleweed Rover \cite{behar_nasajpl_2004} are energy-efficient, making them ideal for remote exploration where energy conservation is critical. Active rolling spherical robots, such as MIT’s Kickbot \cite{batten_kickbot_nodate}, with their low center of gravity and omnidirectional mobility, are robust against external perturbations and excel in tight spaces. However, tumbling robots face challenges. Passive systems often sacrifice controllability for energy efficiency, relying on their shape and external forces for maneuvering. Additionally, rolling robots typically use their entire body for locomotion, leaving no stable platform for sensors, which complicates tasks like localization and perception. 

Early rolling robots like Rollo \cite{halme_motion_1996} and the Spherical Mobile Robot (SMR) \cite{reina_rough-terrain_2004} used a spring-mass system with a driving wheel to create a mass imbalance for movement, but they were unreliable, as the driving wheel often lost contact with the sphere. Furthermore, significant central weight was required to generate enough inertia for propulsion. Other notable designs include the University of Pisa’s Sphericle \cite{bicchi_introducing_1997} and Spider Rolling Robot (SRR) \cite{western_golden_2023}. Sphericle used a car inside a sphere to drive the structure, but relied on gravity to keep the car wheels in contact with the inside of the sphere. Large perturbations led by mobility on rough terrain could dislodge the car and incapacitate the robot. 

A more precise tumbling method involves shifting internal weights to control the center of mass, as seen in the University of Michigan’s Spherobot \cite{mukherjee_simple_1999} and University of Tehran’s August Robot \cite{javadi_a_introducing_2004}, though these systems are not energy-efficient due to the added weight required. 

A more energy-efficient means of positioning the center of gravity for tumbling is by using deformable structures. Successful examples \cite{tian_dynamic_2015, wei_design_2019,wang_trajectory_2018, wang_dynamics_2018,sastra_dynamic_2009} can be identified that have attempted rolling by articulated structural designs that allow such deformation. Notable examples are Ritsumeikan University's Deformable Robot \cite{sugiyama_crawling_2006}, and Ourobot \cite{paskarbeit_ourobotsensorized_2021} with an articulated closed-loop structure.

NASA's Hedgehog \cite{httpswwwjplnasagov_hedgehog_nodate} 
is a notable example developed specifically for space applications, derived from ETH's \textit{Cubli} \cite{gajamohan_cubli_2012}. It combines hopping, oblique hopping, tumbling, escape motion, and yawing (pointing) to traverse asteroids and comets. While it offers robust locomotion on a flat surface, because of its simple operation principle, Hedgehog has a few serious shortcomings that make it unsuitable for long-distance mobility. First, it can get stuck in soft terrain due to its jagged shape (a cube with protusions at its corners). A tornado (fast rotations around body axes) maneuver is proposed by its designers to escape; however, any fast body movements in these conditions can aggravate the situation. Next, because of its cubic shape, Hedgehog's tumbling can be rough compared to other smoother circular geometries. Last, Hedgehog's actuation relies on spinning three internal flywheels with considerable inertia to produce reaction torque. This actuation can be very costly, particularly for larger versions of Hedgehog. Therefore, if large-payload systems are considered in future NASA Moon missions, Hedgehog may suffer from scalability issues.

\subsection{Proposed Solution and Design Rationale}
Our proposed system, COBRA, is a snake-inspired multimodal rover designed for challenging terrain exploration (Fig. \ref{fig:cobra}). COBRA combines the advantages of tumbling locomotion with the morphology afforded by snake robots to effectively address the locomotion challenges of steep crater walls as well as the uneven terrain of the crater floor. COBRA is lightweight at just 7.11 kg, with a compact diameter of 10 cm and a length of 1.7 m. Its eleven actuated degrees of freedom (DOF) enable it to morph its body shape to achieve various locomotion modes that can adapt to unpredictable and rough terrain. To achieve tumbling locomotion, COBRA raises and links its head and tail modules together to transform into a wheel-like structure. We identify these two distinct configurations as (1) Snake Configuration, where the head and tail is unconnected, and (2) Hex-Ring Configuration, where the head and tail are connected to form a closed loop.

In Snake Configuration, COBRA employs sidewinding and other slithering gaits to move efficiently across flat or uphill terrain. Sidewinding is a gait used by snakes to traverse loose or slippery surfaces like sand, and is particularly relevant for traversing lunar regolith, which shares similar properties. Variations of sidewinding locomotion have been shown to be fast and efficient, able to push off rocks and other obstacles to minimize energy consumption and increase traction \cite{wang_directional_2020, sanfilippo_perception-driven_2017, holden_optimal_2014}. Further, the symmetric snake-like morphology makes it excellent at traversing uneven and unknown terrain by virtue of not needing to reorient itself. The system's weight is distributed along the entire length of its body which mitigates sinkage, and also allows for a large payload capacity, which can carry sensors such as a spectrometer to determine the concentration of hydrogen in the lunar regolith. By swinging its joints back and forth, the robot can also dig into the ground, exposing scientific samples below the surface or clearing material away if the system is stuck. It can also lift sections of its body into the air to overcome obstacles, climb out of holes, and aim instruments for navigation or communication.

For tumbling down slopes, COBRA enters the Ring Configuration and shifts its center of gravity to initiate tumbling. Through manipulation of its posture within the ring configuration using its joint actuators, COBRA can actively steer itself with minimal effort, allowing it to track desired paths to reach points of interest along the slope or avoid obstacles. These capabilities are not limited to lunar exploration. As a ground based system, it is a powerful tool for exploration of similar environments on Mars, such as Valles Marineris, a canyon system that features large, sloped terrain that our system can efficiently travel down to perform in-situ measurements in channels that potentially contain water.

Some examples of tumbling platforms employing active deformation have been discussed above. COBRA distinguishes itself from these platforms in two significant ways: (1) \textit{Dynamic Tumbling Locomotion:} As a tumbling system, it is field-tested and capable of dynamic tumbling locomotion with steering in two dimensions (2) \textit{Multimodal Locomotion and Manipulation:} COBRA is a multimodal robot that employs the Snake Configuration to perform various locomotion and manipulation tasks on flat ground. This feature extends its operational scope beyond that of typical tumbling robots, enabling it to address diverse mission requirements.

Previous studies have extensively explored snake locomotion across various robotic platforms \cite{rollinson_design_2014, liljeback_mamba_2014, ramesh_sensnake_2022, seeja_survey_2022, transeth_survey_2009, marvi_sidewinding_2014}. COBRA builds on these works by implementing a broad range of snake gaits, achieving untethered operation, and integrating these capabilities with tumbling locomotion. This enables COBRA to operate effectively across a wide range of environments of various slopes and roughness, making it highly suitable for the exploratory nature of the proposed task. 

A comparable platform of note is NASA JPL's EELS snake robot \cite{vaquero_eels_2024}, which employs a screw propulsion mechanism along the length of its body, enabling locomotion on low-friction ice surfaces such as those found on Europa.  However, this mechanism is less suited for granular and rocky terrains like those encountered on the Moon. COBRA is designed with locomotion strategies tailored for high-roughness surfaces, utilizing slithering and sidewinding gaits for flat or slightly inclined terrain and tumbling locomotion for rapid movement across steep slopes. This combination makes COBRA uniquely suited for environments where rapid elevation changes and substrate variability are critical considerations, for example lunar sites such as Shackleton Crater.



\subsection{Envisioned Mission Scenario}
\label{sec:mission}

\begin{figure}
    \centering
    \includegraphics[width=1.0\linewidth]{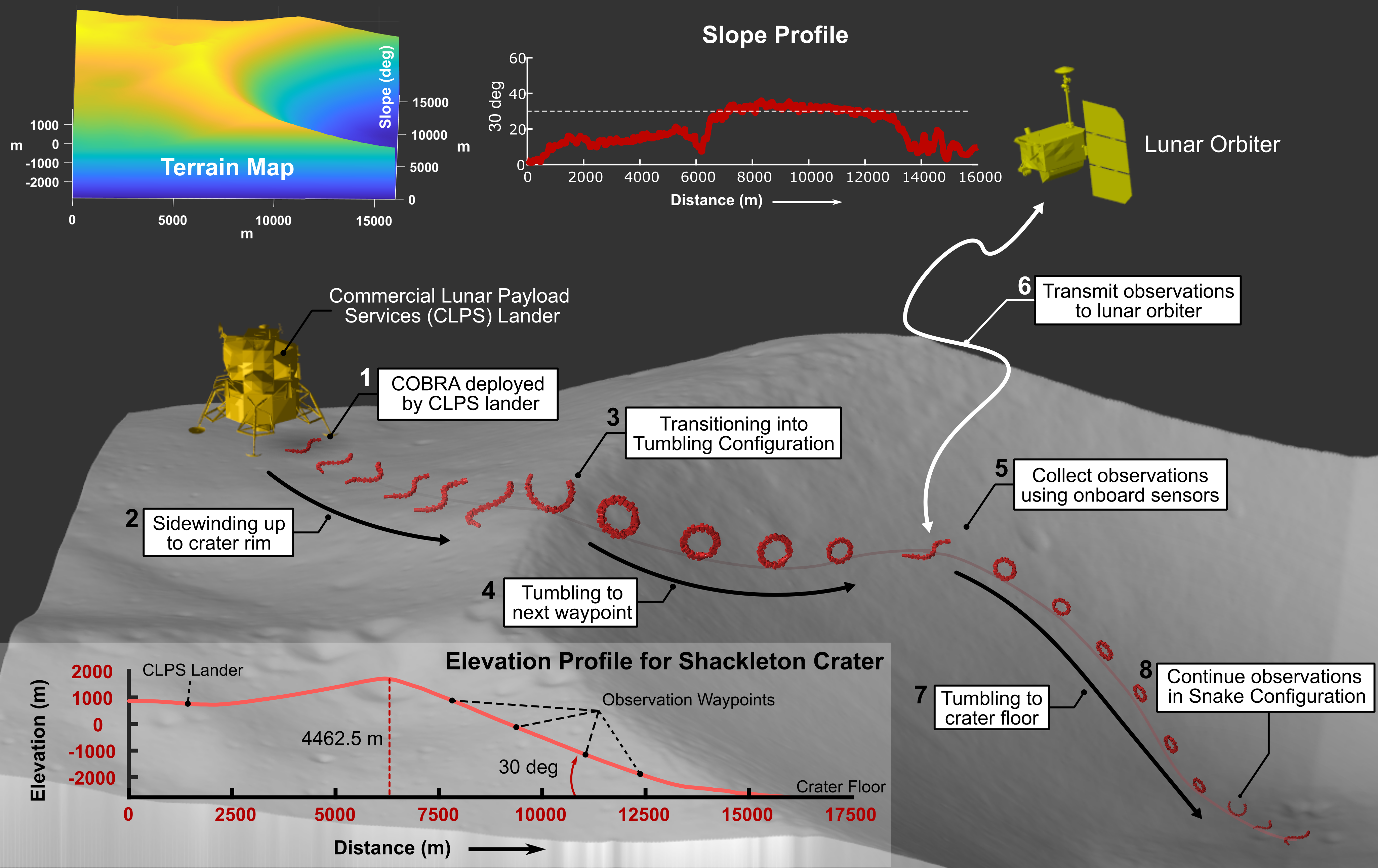}
    \caption{%
\modify{\textbf{Envisioned COBRA mission scenario within Shackleton Crater at the lunar south pole.}%
~Following a precision landing via the Commercial Lunar Payload Services (CLPS) initiative on a sunlit ridge near the crater rim \textbf{(1)}, the bioinspired COBRA robot is deployed to begin its descent mission. It initiates traversal using sidewinding locomotion across the relatively flat regolith to reach the break-in-slope at the crater's edge \textbf{(2)}. Upon reaching the slope, COBRA reconfigures its body into a mechanically interlocked hexagonal ring, termed the \emph{hex-ring} mode \textbf{(3)}. This reconfiguration enables controlled tumbling descent by shifting its center of mass and exploiting gravitational potential on the inclined terrain. During the descent down the $30^\circ$ crater wall, COBRA periodically halts at predesignated locations. At each of these observation nodes, the robot transitions back into its elongated configuration to perform scientific measurements (e.g., neutron and near-infrared spectroscopy) and establish high-bandwidth communication links with a polar orbiting relay for data transfer \textbf{(4--6)}. After reaching the crater floor via a final passive tumble \textbf{(7)}, COBRA resumes sidewinding locomotion \textbf{(8)} across the permanently shadowed region. In this phase, it continues mapping hydrogen signatures and terrain microstructure. \textbf{Top and bottom:} Terrain slope map of the crater interior, visualized as a 2D colormap. Warmer hues (yellow-orange) represent shallower gradients near the crater rim, while cooler hues (green-blue) indicate steeper slopes approaching and exceeding approx. $30^\circ$ in the inner wall region. The slope increases markedly after approx. $7500\;\mathrm{m}$ from the landing site, indicating the start of the break-in-slope. The longitudinal profile confirms sustained slopes of $30^\circ$--$40^\circ$ over a span of approx. 5 km before gradually reducing near the talus and floor region. This terrain is well-suited for controlled tumbling enabled by the robot’s reconfigurable morphology.%
}}
\label{fig:mission-scenario}
\end{figure}

Here we present the mission scenario envisioned for COBRA’s operation considering the case study of Shackleton Crater. COBRA will be deployed from a Commercial Lunar Payload Services (CLPS) lander near the edge of Shackleton Crater, as illustrated in Fig.~\ref{fig:mission-scenario}. Following deployment, the system will navigate to the slope of the crater using sidewinding locomotion. In this mode, segments of the robot's body are lifted and shifted forward, while others remain in contact with the ground, minimizing shear forces at contact points and preventing slippage on the soft regolith.

Next, COBRA leverages the Moon's partial gravity to tumble down the steep slope in its wheel-like configuration. Transitioning from the snake configuration, the system shifts its internal weight onto the slope to initiate tumbling. During descent, the joints remain static, conserving energy.

The system will tumble incrementally, halting every 500 meters to disconnect its head and tail for in-situ measurements. The scientific payload, a spectrometer housed in the tail, can be positioned using COBRA’s joints. This allows the system to create a detailed hydrogen concentration map at various depths within the crater. After collecting data, COBRA transmits the results to a lunar orbiter via a radio antenna in its head module, eliminating the need for the system to climb out of the crater for data transmission.

Following communication at the sampling location, COBRA reforms its tumbling structure, tumbles another 500 meters, and repeats the process. Upon reaching the flatter center of Shackleton Crater, it will switch to sidewinding to continue its exploration and data collection until its power is exhausted.

Based on this mission scenario, we come up with four objectives that we focus on for this work. 1) Locomotion over flat or slightly sloped ground using snake-like gaits. 2) Locomotion down slopes using two tumbling configurations. 3) Multi-modal operation allowing seamless switching between the modes of locomotion with no human contact with the robot. 4) Modular design that is robust to potential failures that can occur during remote operation in a space mission.

For this prototype, we focused on developing COBRA's locomotion capabilities and designing a robust, modular system. While not using space-grade materials, the prototype is built to withstand outdoor terrestrial conditions. The following sections detail our development and testing process towards achieving these objectives.


\section{Overview of Hardware Design}

\begin{figure}
    \centering
    \includegraphics[width=0.9\textwidth]{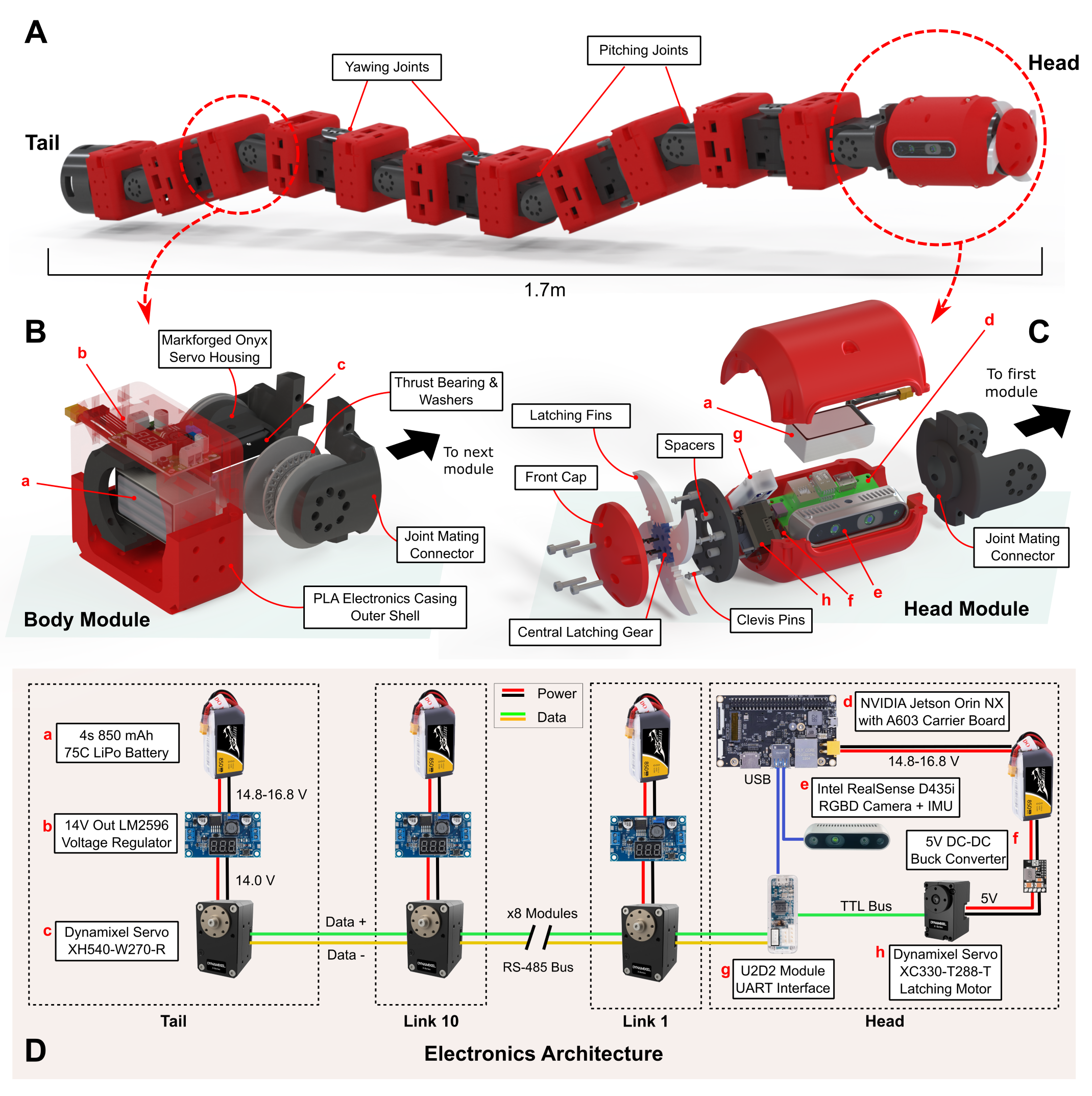}
    \caption{\modify{\textbf{Overview of the COBRA robotic platform and its internal subsystems.}%
    ~\textbf{(A)}~Fully assembled COBRA prototype designed by Northeastern University. The platform features a modular, snake-inspired design that supports both sidewinding locomotion and reconfiguration into a closed hexagonal ring for tumbling descent. The robot consists of 10 identical articulated body modules and two specialized end modules (head and tail).%
    ~\textbf{(B)}~Interior view of a representative body module, showing the core electromechanical components: \textbf{(a)} a lithium-polymer battery pack (Tattu 850 mAh 4s 150c) for local power, \textbf{(b)} an onboard voltage regulator (LM2596 Variable Voltage Converter) for power regulation, and \textbf{(c)} a high-torque smart servo actuator (Dynamixel XH540-W270-R) for joint articulation. Each module operates semi-independently, enabling distributed control and robustness.%
    ~\textbf{(C)}~Internal layout of the head module, which integrates \textbf{(d)} the robot's central processing unit (NVidia Jetson Orin NX 8GB), \textbf{(e)} inertial measurement unit (IMU) and stereo camera (Intel RealSense D435i), \textbf{(a)} battery (Tattu 850 mAh 4s 150c), \textbf{(h)} and servo for claw actuation (Dynamixel XC330-T288-T). These components support state estimation, localization, and autonomy during both sidewinding and tumbling phases.%
    ~\textbf{(D)}~System architecture diagram depicting key electrical and communication pathways. Data and control signals are routed between modules via an I\textsuperscript{2}C-based backbone, while the head module coordinates global decision-making. The modular design supports scalability, distributed computing, and fault tolerance critical for operations in extreme lunar terrain.%
}}
    \label{fig:cobra}
\end{figure}

COBRA consists of eleven single-degree-of-freedom actuated joints (Fig. \ref{fig:cobra}-A), alternating between pitching and yawing directions. The front of the robot consists of a head module (Fig. \ref{fig:cobra}-C) containing the computing and perception sensors. At the end, there is an interchangeable payload module that can house various scientific instruments. For our mission, we consider a mass spectrometer. The rest of the system consists of identical 1-DOF joint modules that contain joint actuators and batteries (Fig. \ref{fig:cobra}-B). The following sections detail the design features of each module and their contributions to COBRA's overall functionality.

\subsection{Body Module} 
COBRA features a modular design comprising of ten identical and interchangeable body modules situated between the head and tail. This modularity provides significant advantages in terms of operation, design, and manufacturability, including streamlined prototyping and assembly. As illustrated in Fig.~\ref{fig:cobra}-B, each body module includes a 1-DOF joint with a 150-degree range of motion. The modules are connected using circular male and female connectors located at their front and rear, enabling straightforward replacement or extension without altering system functionality.

Each module is powered by a DYNAMIXEL XH-540-W270-R servo motor \cite{dynamixel}, which provides precise position control via absolute encoders. The servos are linked through an RS-485 daisy-chain communication system, with unique IDs assigned to each motor, ensuring continuous operation even if individual servos fail. The entire system is managed by an NVIDIA Jetson Orin NX computer housed in the head module.

The power and communication architecture of COBRA, shown in Fig.~\ref{fig:cobra}-D, is designed to evenly distribute power across the system. Each body module contains an independent battery connected to the motor through a voltage regulator, simplifying power distribution and allowing modules to be added or removed without impacting functionality. This decentralized setup enhances system resilience to failures. GensTattu 850mAh 14.8V LiPo batteries, chosen for their lightweight (107 g) and compact design, provide an operational runtime of 2 hours under typical conditions (0.5 A average current).

The joint housing for each module is fabricated through 3D printing using a Markforged printer with carbon fiber inlay, yielding lightweight, durable components, designed to withstand the torque requirements of the tumbling transformation (7 Nm peak for the terrestrial prototype) and incorporating a safety margin of 2x for impact forces during tumbling, as these represent the system's peak torque requirements.

Each module is enclosed in a 3D-printed PLA outer shell, which serves dual purposes. First, it protects internal electronics, including the battery and voltage regulator, from external impacts. Second, the shell shapes the robot's geometry for tumbling locomotion. Its flat bottom provides a stable base for transitions into the tumbling configuration and aids in stability during tumbling, while maintaining the compact footprint of the robot.

\subsection{Head Module and Latching Mechanism} 
COBRA's head module, shown in Fig.~\ref{fig:cobra}-C, houses an NVIDIA Jetson Orin NX as the primary processor, an Intel RealSense D435i Stereo Camera with an IMU for navigation, and an RS-485 to UART interface module to enable communication with the joint servos. These are powered by the same 850 mAh LiPo battery used in the body modules. The housing of the head module is 3D-printed with PLA and the overall weight matches the body modules (600 g) to keep the mass distribution uniform. 

To enable tumbling locomotion, the head module features a latching mechanism which allows the robot to lock the head and tail together, forming a wheel-like structure. The mechanism includes four latching fins that sit flush with the surface of the head module when closed and are driven by a central gear by a Dynamixel XC330-T288-T servo to open and engage with matching slots in the tail module. This creates a rigid connection that requires minimal energy to maintain and can withstand the forces generated during tumbling.

The choice for an active latching mechanism design stemmed from the design requirements and restrictions. Magnets were initially discussed as a passive latching option, however they would not be effective in conjunction with the ferromagnetic regolith. Further, due to the need to maintain a latched configuration even when a large amount of force is applied to the system during tumbling, a passive system was not chosen, for there would be an increased risk of unlatching during tumbling. The chosen design connects the head and tail modules rigidly (except for minor tolerance) and minimizes the possibility of latching failure during tumbling. Shear load on the latching fins was evaluated through simulations and different configurations of carbon fiber reinforcement layers were tested within the 3D-printed parts to meet the requirements. The final configuration consists of four 6-mm-thick fins with three equally spaced layers of carbon fiber reinforcement. The latching mechanism is enclosed by a front cap that mitigates dust infiltration when the fins are closed. This can be further dust-proofed in the open configuration by using a flexible membrane that encloses the fins and stretches with them when open to keep the mechanism protected.

\subsection{Tail Module} 
The tail module includes an interchangeable payload section and slots for the latching fins to engage during tumbling. The payload section is designed to accommodate scientific instruments needed for exploration and analysis, such as a spectrometer for detecting water ice. It can also house mapping tools like radars or environmental monitoring sensors. Leveraging COBRA's articulated body, the robot can integrate locomotion and manipulation to position and orient these sensors as needed for effective data collection and analysis.

\section{Modeling and Gait Design}\
\label{sec:controls}

\begin{figure}
    \centering
    \includegraphics[width=0.9\linewidth]{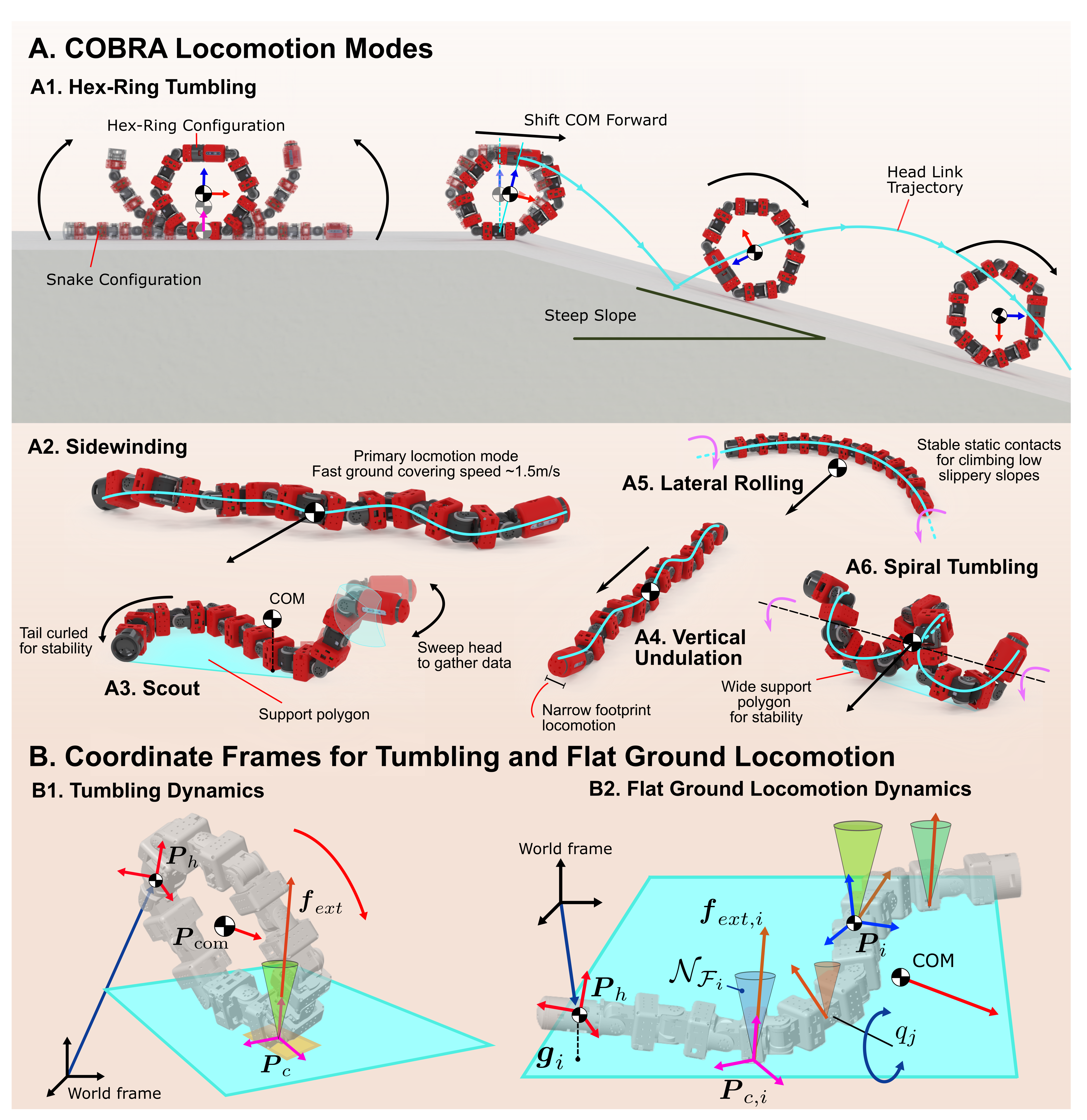}
    \caption{%
    \modify{\textbf{COBRA's multi‐modal locomotion capabilities and dynamic modeling schematic.}
    \textbf{(A)} presents six distinct locomotion behaviors enabled by COBRA's reconfigurable morphology: 
    \emph{Hex‐Ring Tumbling} \textbf{(A1)} leverages the head-tail linkage to form a rigid ring that initiates controlled descents down steep slopes with minimal energy expenditure; 
    \emph{Sidewinding} \textbf{(A2) }generates orthogonal sinusoidal waves in the body to traverse flat or irregular terrain at speeds up to 1.5 m/s; 
    \emph{Scout Mode} \textbf{(A3)} actuates joints to raise and pivot the head module for panoramic terrain mapping and in‐situ scientific measurements; 
    \emph{Vertical Undulation} \textbf{(A4)} propagates bending waves along the robot's longitudinal axis to navigate narrow passages comparable to its 10 cm diameter; 
    \emph{Lateral Rolling} \textbf{(A5)} alternates static contacts and torsional body twists to climb slippery uphill slopes with enhanced stability; 
    and \emph{Spiral Tumbling} \textbf{(A6)} reconfigures into a helical structure to maintain balance while tumbling on uneven, low-angle surfaces.  
    \textbf{(B)} shows the free‐body and contact‐force schematic used in the dynamic modeling: an inertial (black), head (red), local link (blue), and contact frames (magenta) along the body. Ground reaction forces $\mathrm{f_{ext,i}}\in \mathcal{N}_{\mathcal{F}_i}$ at each contact $\mathrm{P_{c,i}}$ are resolved into normal ($\mathrm{f_{N,i}}$) and tangential ($\mathrm{f_{T,i}}$) components and mapped into generalized coordinates $\mathrm{q_j}$ via the Jacobians $\mathrm{J_i(q)}$, yielding the external force vector 
    $\mathrm{u_{ext}}$ in the Euler-Lagrange equations of motion.%
    }}

    \label{fig:locomotion-modes}
\end{figure}

In Snake Configuration, COBRA relies on contact rich environment interaction to generate movement. In Hex-Ring Configuration, COBRA uses its articulated joints to manipulate its mass distribution to generate angular velocity in the direction of desired steering. In this section, we discuss the modeling and control strategies developed for both locomotion modes.

Using its articulated nature, COBRA can achieve several different types of locomotion useful in various scenario (Fig. \ref{fig:locomotion-modes}A). Hex-Ring tumbling locomotion (Fig \ref{fig:locomotion-modes}A1) as previously discussed enables the robot to traverse down steep slopes using minimal energy. COBRA can manipulate its joints to steer the robot while in this configuration to reach targets of interest. Sidewinding (Fig. \ref{fig:locomotion-modes}A2) is a fast and efficient form of snake locomotion that can travel up slippery slopes and achieve flat ground locomotion of up to 1.5 m/s. This can be adapted to use the terrain to its advantage and traverse rough surfaces with relative ease. The Scout Mode (Fig. \ref{fig:locomotion-modes}A3) enables COBRA to raise up and sweep its head to scan the terrain for topology mapping or orient sensors and collect scientific data. Vertical Undulation (Fig. \ref{fig:locomotion-modes}A4) generates motion along the length of the body, enabling COBRA to travel through narrow gaps comparable in size to the diameter of the body (10 cm). The Lateral Rolling gait (Fig. \ref{fig:locomotion-modes}A5) has static contacts, twisting the body about its axis to step forward while maintaining a stable base. This has been used to travel uphill on slopes that other gaits such as sidewinding might slip on. Spiral tumbling (Fig. \ref{fig:locomotion-modes}A6) is a variation on Hex-Ring tumbling that utilizes a helical structure to provide more stability at the cost of tumbling speed. This may be used on shallower, uneven slopes where the Hex-Ring Configuration would struggle to maintain balance to tumble effectively.

In the following section, we discuss the modeling and optimal control-based approach proposed for controlling snake-like locomotion gaits.


\subsection{Offline Contact-Implicit Gait Optimization}
\subsubsection{Motion Parameterization}
Figure. \ref{fig:locomotion-modes}B2 depicts a general snapshot of COBRA performing a snake-gait locomotion. This mode is characterized by an open kinematic chain (see Supporting Information for derivations) from the head to the tail with multiple contacts with the ground along the body, represented by $\bm P_{c,i}$ for the $i^\text{th}$ contact point. For a 12 link robot, this can be anywhere from 2 to 12 contacts. We model this as a constrained optimization problem to minimize overall energy of the system while solving for ground contact forces and joint torques to achieve a desired center of mass acceleration. 

To derive the generalized equations of motion for the robot, we assume that the pose of the head link is known through state estimation techniques and resolved as a 3-DOF position vector 
\[
\bm P_h = [x_h,~y_h,~z_h]^\top
\] 
and Euler angle orientations 
\[
\bm \Phi_h = [q_r,~q_p,~q_y]^\top.
\]
The 6-DOF head pose is then represented by $\bm q_H = [\bm P_h^\top,~\bm \Phi_h^\top]^\top$. The joint angles $\bm q_J = [q_1,~\dots,~q_N]^\top$, where $N$ is the number of joints in the robot, are known through continuous sampling of absolute encoders in each joint actuators. The generalized positions for the robot can then be written as 
\[
\bm q = [\bm q_H^\top,~\bm q_J^\top]^\top,\]
with the state vector for the robot represented as: 
$\bm x = [\bm q^\top,~ \bm {\dot q}^\top]^\top$. The dynamical equations of motion are derived from the Euler-Lagrange Formalism after obtaining the system's Lagrangian $\mathcal{L}$:

\begin{equation}
\frac{d}{dt} \frac{\partial \mathcal{L}}{\partial \bm {\dot q}}  - \frac{\partial \mathcal{L}}{\partial \bm q} = \bm u_{ext}
\label{eq:general-EL-eq}    
\end{equation}

where $\bm u_{ext}$ is the sum of generalized forces including joint torques $\bm \tau$ and external ground reaction forces $\bm f_{ext}$. This can then be reduced to the form:

\begin{equation}
    \begin{aligned}
        \bm M(\bm q) \dot{\bm u} - \bm h(\bm q, \bm u, \bm \tau) &= \sum_i \bm J_i^{\top}(\bm q) \bm f_{ext,i},\\
        \bm h(\bm q, \bm u, \bm \tau) = \bm C(\bm q,\bm u)\bm u& + \bm G(\bm q) + \bm B(\bm q)\bm \tau
    \end{aligned}
    \label{eq:eom}
\end{equation}

where $\bm M(\bm q)$ is the mass-inertia matrix, $\bm h(\bm q, \bm u, \bm \tau)$ encompasses centrifugal and Coriolis, $\bm C(\bm q,~\bm u) \bm u$, gravity, $\bm G(\bm q)$, and actuation, $\bm B(\bm q)\bm \tau$, terms. The term $\bm f_{ext,i}$ represents the external forces for the $i^\text{th}$ contact point.

\subsubsection{Contact-Implicit Path Planning}
Next, we formulate the contact implicit optimization following the contact modeling approach in \cite{studer_numerics_2009}. We define gap function $\bm g_i$ that represents the distance between potential contact points on the robot and the terrain. All external forces $\bm f_{ext,i}$ are assumed to stem solely from contact forces between the ground surface and the robot and act unilaterally. 

We can then define the following conditions to indicate that the relative displacement, velocity and acceleration of the body module with respect to the terrain must respect feasible ground reaction forces represented by normal cone inclusion ($\mathcal{N}_{\mathcal{F}_i}(.)$).

\begin{equation}
    \begin{aligned}
\text{1)}~~~~~        -\bm g_i & \in \partial \Psi_{i}\left(\bm f_{ext,i}\right) \equiv \mathcal{N}_{\mathcal{F}_i}\left(\bm f_{ext,i}\right) \\
\text{2)}~~~~~        -\dot{\bm g}_i & \in \partial \Psi_{i}\left(\bm f_{ext,i}\right) \equiv \mathcal{N}_{\mathcal{F}_i}\left(\bm f_{ext,i}\right) \\
\text{3)}~~~~~        -\ddot{\bm g}_i & \in \partial \Psi_{i}\left(\bm f_{ext,i}\right) \equiv \mathcal{N}_{\mathcal{F}_i}\left(\bm f_{ext,i}\right)
    \end{aligned}
    \label{eq:normal-cone-inclusion}
\end{equation}

where $\Psi_i(.)$ denotes the indicator function representing the set of feasible contact forces at the $i^\text{th}$ contact. The gap function $\bm{g}_i$ is defined such that its total time derivative yields the relative constraint velocity $\dot{\bm{g}}_i = \left( \partial \bm{g}_i / \partial \bm q \right)^{\top} \bm u + \partial \bm{g}_i / \partial t$ (the last term captures any explicit time dependency of the terrain geometry, e.g., soft or moving terrain). The time derivative of the relative constraint velocity (see Supporting Information) yields the relative constraint accelerations 
\[
\ddot{\boldsymbol{g}}_i(\boldsymbol{q},t) = \left(\frac{\partial \boldsymbol{g}_i}{\partial \boldsymbol{q}}\right)^\top \dot{\boldsymbol{u}} + \boldsymbol{u}^\top \frac{\partial^2 \boldsymbol{g}_i}{\partial \boldsymbol{q}^2}\boldsymbol{u} + 2\boldsymbol{u}^\top \frac{\partial^2 \boldsymbol{g}_i}{\partial \boldsymbol{q}\partial t} + \frac{\partial^2 \boldsymbol{g}_i}{\partial t^2}
\]
In Eq.~\ref{eq:normal-cone-inclusion}, the normal cone inclusion condition (1) indirectly imposes complementarity constraints on the forces: i) normal forces must be nonnegative (no pulling allowed), ii) friction forces must satisfy Coulomb friction constraints. At velocity and acceleration levels (2-3), this negative sign ensures correct alignment between the contacts' velocity or acceleration and the permissible forces. Hence, we describe a geometric constraint on the acceleration level such that the initial conditions are fulfilled on velocity and displacement levels:
\begin{equation}
    \begin{aligned}
        &\bm g_i(\bm q, t)=0, \\
        &\dot{\bm g}_i=\bm W_i^\top \bm u+\zeta_i=0, \\
        &\ddot{\bm g}_i=\bm W_i^\top \dot{\bm u}+\hat \zeta_i=0,\\
        &\dot{\bm g}_i\left(\bm q_0, \bm u_0, t_0\right)=0,\\
        &\partial \bm{g}_i / \partial t = 0
    \end{aligned}   
\end{equation}
where $\bm W_i = \bm W_i(\bm q, t) = \left( \partial \bm{g}_i / \partial \bm q \right)^{\top}$ and $\zeta_i, \hat \zeta_i$ are residual terms in $\dot{\bm g}_i$ and $\ddot{\bm g}_i$. This implies that the generalized constraint forces must be perpendicular to the manifolds $\bm g_i=0$, $\dot{\bm g_i}=0$, and $\ddot{\bm g}_i=0$. 

We divide the contact forces into normal and tangential components, denoted as 
\[
\bm f_{ext,i}=\left[\begin{array}{ll}f_{N,i}, & \bm f_{T,i}^{\top}\end{array}\right]^{\top} \in \mathcal{F}_{i}.
\] 
In this context, the force space $\mathcal{F}_i$ facilitates the specification of non-negative normal forces ($\mathbb{R}_{0}^+$) and tangential forces adhering to Coulomb friction $\left\{\bm f_{T,i} \in \mathbb{R}^2, ||\bm f_{T,i}||<\mu\left|f_{N,i}\right| \right\}$, with $\mu$ representing the friction coefficient.

The underlying rationale behind this approach is that while the force remains confined within the interior of its designated subspace, the contact velocity in the normal direction remains constrained to zero. Conversely, non-zero gap velocities only arise when the forces reach the boundary of their permissible set, indicating either a zero normal force or the maximum friction force opposing the direction of motion.

To formulate these constraints as an optimization problem, it is useful to write the dynamics equation Eq. \ref{eq:eom} using the relationship:

\begin{equation}
    \dot{\bm g}=\bm J_{\mathrm{c}} \bm u,     
\end{equation}
where $\bm J_{\mathrm{c}}$ is the Jacobian matrix. The form explicitly assumes that the terrain is static, i.e., \(
\frac{\partial \boldsymbol{g}}{\partial t}=0,\) and that the individual gradients \( \boldsymbol{W}_i \) are combined into a single comprehensive Jacobian \(\boldsymbol{J}_c\). Recall from Eq.~\ref{eq:eom}, the generalized external forces are given by \(
\sum_{i}\boldsymbol{J}_{i}^{\top}(\boldsymbol{q})\boldsymbol{f}_{\text{ext},i}\). Here, \(\boldsymbol{J}_{i}(\boldsymbol{q})\) explicitly relates external contact forces at contact \(i\) to generalized forces (joint-space). It provides a mapping between external forces at point \(i\) and generalized coordinates.

In contrast, the Jacobian \(\boldsymbol{J}_{c}\) is derived directly from \(\boldsymbol{J}_{i}\). Specifically, for the contact point \(i\), if we define the normal direction \(\boldsymbol{n}_{i}\), we have:
\[
\boldsymbol{W}_{i}^{\top} = \boldsymbol{n}_{i}^{\top}\boldsymbol{J}_{i}
\]
where \(\boldsymbol{n}_{i}\) is the unit vector normal to the terrain at the contact point \(i\). Thus, explicitly, the full contact Jacobian \(\boldsymbol{J}_{c}\) can be written in terms of \(\bm J_i\):
\[
\boldsymbol{J}_{c} = 
\begin{bmatrix}
\boldsymbol{n}_{1}^{\top}\boldsymbol{J}_{1} \\
\boldsymbol{n}_{2}^{\top}\boldsymbol{J}_{2}\\
\vdots\\
\boldsymbol{n}_{n}^{\top}\boldsymbol{J}_{n}
\end{bmatrix},
\]
Each row of \(\boldsymbol{J}_{c}\) thus represents the projection of the spatial velocity at contact points onto the terrain’s normal directions, explicitly connecting joint-space velocities to gap velocities.

Differentiating the gap vector function with respect to time and substituting in Eq. \ref{eq:eom} gives:

\begin{equation}
    \ddot{\bm g} = \bm J_c \bm M^{-1} \bm J_c^{\top} \bm f_{ext} + \dot{\bm J}_c \bm u + \bm J_c \bm M^{-1} \bm h,
\end{equation}

where $\bm G = \bm J_c \bm M^{-1} \bm J_c^{\top}$, the Delassus matrix, signifies the apparent inverse inertia at the contact points (or, shows how external forces explicitly influence gap accelerations via the inertia of the robot), $\dot{\bm J}_c \bm u$ accounts explicitly for changing geometry (time-varying Jacobian), and $\bm J_c \bm M^{-1} \bm h$ encapsulates all terms independent of the stacked external forces $\bm f_{ext}$, i.e., contributions from actuation torques and nonlinear terms (gravity, centrifugal, Coriolis). 

We can now define the constrained optimization problem according to Gauss principle of least constraint such that the constrained and free systems behave as close as possible (see Supporting Information):
\[
\begin{aligned}
&\underset{\boldsymbol{q}(t),\boldsymbol{u}(t),\dot{\boldsymbol{u}}(t),\boldsymbol{\tau}(t),\boldsymbol{f}_{\text{ext},i}(t)}{\text{minimize}} 
 \frac{1}{2}\,\boldsymbol{f}_{\text{ext}}^\top\,\boldsymbol{G}\,\boldsymbol{f}_{\text{ext}}
\;,\\
&\text{subject to:}\\
&(1) \quad \boldsymbol{M}(\boldsymbol{q})\,\dot{\boldsymbol{u}} -\boldsymbol{h}(\boldsymbol{q},\boldsymbol{u},\boldsymbol{\tau})
= \sum_{i=1}^m \boldsymbol{J}_i^\top(\boldsymbol{q})\,\boldsymbol{f}_{\text{ext},i},\\
&(2) \quad \boldsymbol{q}_{\min} \leq \boldsymbol{q}(t) \leq \boldsymbol{q}_{\max},\\
&(3) \quad \boldsymbol{\tau}_{\min} \leq \boldsymbol{\tau}(t) \leq \boldsymbol{\tau}_{\max},\\
&(4) \quad \bm g_i(\boldsymbol{q})\geq 0,\quad f_{N,i} \geq 0,\quad \bm g_i(\boldsymbol{q}) f_{N,i} = 0,\\
&(5) \quad \|\boldsymbol{f}_{T,i}\| \leq \mu_i f_{N,i},\quad 
f_{N,i} = \boldsymbol{n}_i^\top \boldsymbol{f}_{\text{ext},i},\quad
\boldsymbol{f}_{T,i} = \boldsymbol{f}_{\text{ext},i} - \boldsymbol{n}_i (\boldsymbol{n}_i^\top \boldsymbol{f}_{\text{ext},i}),\\
&(6) \quad \dot{\bm g}_i=\bm W_i^\top \bm u+\zeta_i=0, \\
&(7) \quad \ddot{\bm g}_i=\bm W_i^\top \dot{\bm u}+\hat \zeta_i=0,\\
&(8) \quad \dot{\bm g}_i\left(\bm q_0, \bm u_0, t_0\right)=0,\\
%
\end{aligned}
\] 
In the contact-implicit optimization problem, 1-6 denote dynamics, joint motion, actuation range, complementarity, second-order cone (SOC), and contact velocity constraints, respectively. Solving this optimization using a shooting method returns predicted external forces and joint inputs to achieve a desired robot acceleration.

\subsection{Mixed Hamiltonian-Lagrangian Model of Tumbling}

Tumbling locomotion is executed in two stages. First the robot transforms from its snake configuration to its wheel-like configuration by raising and docking the head and tail modules into a hexagonal structure. Due to the accurate position tracking by the actuators, this movement can be repeatably performed in an open-loop fashion. There is a design provision that facilitates this process. The domed cap of the head module, shown in Fig. \ref{fig:cobra}C, helps guide the head into the tail module for minor position offsets due to compliance of the structure. Once the head and tail module are concentrically aligned, the latching fins extend to lock them in place, completing the transformation.

The second stage is the tumbling phase. Once the robot starts tumbling, it retains a symmetric hexagonal structure and uses minimum-energy tumbling. If the robot topples over during tumbling, it is able to return to its snake configuration, reorient itself and continue tumbling, adding a layer of robustness to this mode of locomotion. In this work, the tumbling structure is passive. However, it is possible to manipulate tumbling heading angle through posture manipulation \cite{salagame_dynamic_2024} which is overlooked in this paper. We only consider rolling speed manipulations using shape control as is outlined below. 

To model the tumbling dynamics, we use Eq.~\ref{eq:eom}. The forces and coordinate frames used here are shown in Fig. \ref{fig:locomotion-modes}-B1. The symmetry sources in the dynamics, i.e., cyclic variables, including the head roll, pitch and yaw angles, motivate model partitioning (see Supporting Information) and re-writing Eq.~\ref{eq:eom} in a block-matrix, state-space form:
\[
\begin{aligned}
\begin{bmatrix}
\dot{\boldsymbol{q}}_J \\
\dot{\boldsymbol{u}}_J \\
\dot{\boldsymbol{q}}_H \\
\dot{\boldsymbol{\sigma}}_{\text{cm}}
\end{bmatrix}
&=
\begin{bmatrix}
\boldsymbol{0} & \boldsymbol{I} & \boldsymbol{0} & \boldsymbol{0} \\
\boldsymbol{0} & -\bm M'^{-1}\boldsymbol{M}_{JH}\boldsymbol{M}_{HH}^{-1}\boldsymbol{M}_{HJ} & \boldsymbol{0} & -\bm M'^{-1}\boldsymbol{M}_{JH}\boldsymbol{M}_{HH}^{-1} \\
\boldsymbol{0} & -\boldsymbol{M}_{HH}^{-1}\boldsymbol{M}_{HJ} & \boldsymbol{0} & \boldsymbol{M}_{HH}^{-1} \\
\boldsymbol{0} & \boldsymbol{0} & \boldsymbol{0} & \boldsymbol{0}
\end{bmatrix}
\begin{bmatrix}
\boldsymbol{q}_J \\
\boldsymbol{u}_J \\
\boldsymbol{q}_H \\
\boldsymbol{\sigma}_{\text{cm}}
\end{bmatrix}
\\&+ 
\begin{bmatrix}
\boldsymbol{0} \\
\bm M'^{-1}\sum_i \boldsymbol{J}_{J,i}^\top \\
\boldsymbol{0} \\
\sum_i \boldsymbol{\Upsilon}_i \times
\end{bmatrix}\boldsymbol{f}_{\text{ext}}+
\begin{bmatrix}
\boldsymbol{0} \\
-\bm M'^{-1}\boldsymbol{h}_J \\
\boldsymbol{0} \\
\boldsymbol{0}
\end{bmatrix}
\end{aligned}
\]
where $\bm{\sigma}_{cm}$ denotes the conjugate momentum to $\bm{q}_H$ (angular momentum about the COM), and $\bm{M}' = (\boldsymbol{M}_{JJ}-\boldsymbol{M}_{JH}\boldsymbol{M}_{HH}^{-1}\boldsymbol{M}_{HJ})$. As the state-space model suggests, $\bm{h}_J$, which embodies joint actions, can be leveraged to steer the posture dynamics $\bm{q}_J$ and $\bm{u}_J$. Then, the posture can be used to control the angular momentum of the tumbling robot. However, we assume the posture is fixed, and the model is purely employed for the dynamical analysis of tumbling.

\section{Results and Discussion}
\subsection{Simulations}

\begin{figure}
    \centering
    \includegraphics[width=1\linewidth]{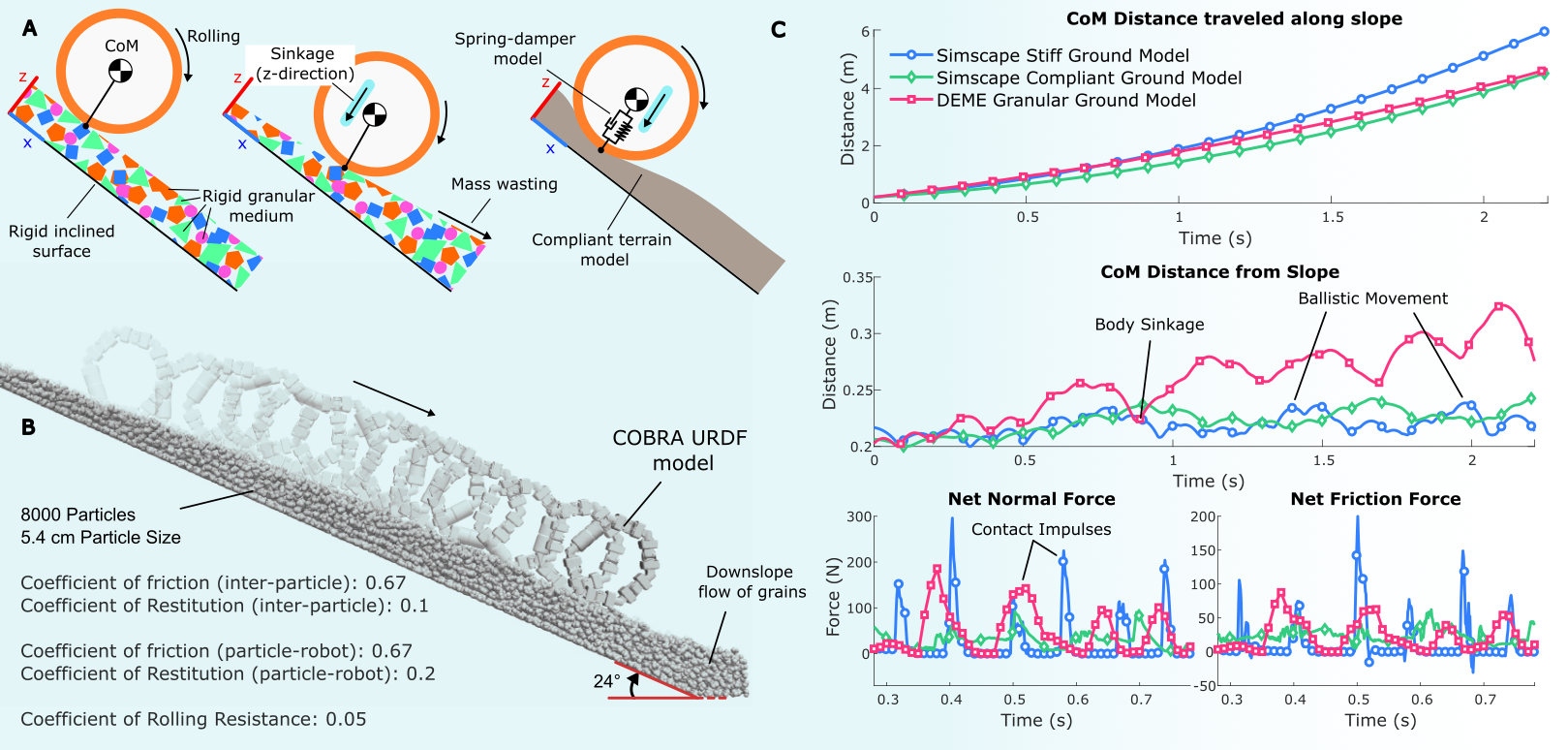}
    \caption{
    \modify{\textbf{Comparison of terrain interaction models during downhill tumbling.} 
    \textbf{(A)} Conceptual diagrams showing two terrain types: a rigid granular medium (left/center), and a compliant deformable terrain (right). Capturing key dynamics such as CoM sinkage, traction, slip, and mass wasting are the goals. 
    \textbf{(B)} Snapshot from the simulated COBRA tumbling over a granular slope composed of 8,000 rigid spherical particles, each 5.4~cm in diameter, based on Discrete Element Method (DEM), using the DEM Engine from Project Chrono \cite{zhang_2024_deme}. The slope is inclined at an angle of 24$^\circ$, and the gravitational acceleration vector is set to [0, 0, -9.81]~m/s$^2$. The coefficient of restitution is 0.1 for inter-particle contacts and 0.2 for particle-robot interactions. In the tangential direction, we use friction coefficients of 0.67 for both inter-particle and particle-robot. Rolling resistance is included with a coefficient of 0.05. Robot is tumbling from rest under the influence of gravity. The ground contact forces and motion dynamics are computed at a simulation time step of 0.5~ms, using a GPU solver, executed on a Nvidia RTX 4050m with 6~GB of GPU memory and 16~GB of RAM. 2~s of settling time is allowed before the robot began interacting with the terrain. Output data--including CoM position, net contact forces, and frictional impulses--are recorded at a sampling rate of 100~Hz. Visualizations are generated at 10~frames per second for post-processing and analysis. Expected mass wasting effects and down-slope grain flow are visible. 
    \textbf{(C)} Quantitative comparisons across three models: \textit{Model 1 - Simscape stiff ground:} Contact model with spring constant of 1e4~N/m, damping coefficient of 1e3~Ns/m, and transition width of 1e-3~m. \textit{Model 2 - Simscape compliant ground:} Spring constant of $\mathrm{k_n} = 10^3$~N/m, damping coefficient $\mathrm{b_n} = 10^4$~Ns/m, and transition region of 0.3~m. \textit{Model 3 - Chrono DEM granular ground:} Parameters as described in (B). \textbf{Top:} CoM distance traveled along the slope over time. \textbf{Middle:} vertical distance of the CoM from the slope, which is the sum of body sinkage and hex structure CoM shifts due to posture changes. \textbf{Bottom left:} On stiff ground, net normal force exhibits sharp, widely spaced spikes approaching 280 N due to impulse impacts. Granular terrain lowers peaks to roughly 190 N due to energy lost to particle rearrangement. Compliant terrain further attenuates impulses below 120 N via damping. \textbf{Bottom right:} Stiff ground yields the largest friction force, swinging to nearly 180 N with distinct stick–slip episodes synchronized to normal‐force spikes. Granular and compliant terrains dampen these peaks to about 120 N and 70 N as grains mobilize and slide and contact point move, respectively.}}
    \label{fig:granular-tumbling}
\end{figure}

\begin{figure}
    \centering
    \includegraphics[width=0.8\linewidth]{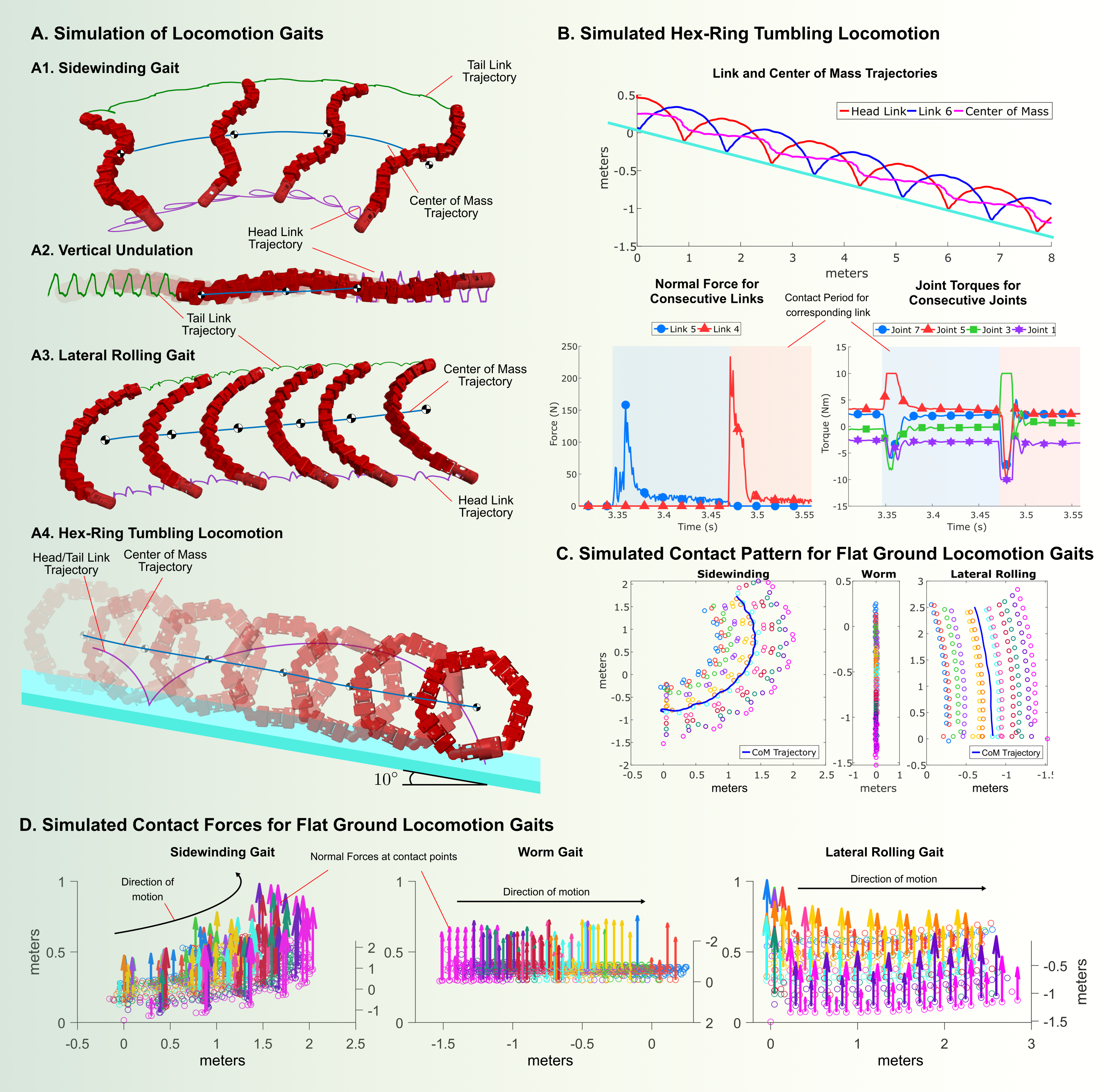}
\caption{%
    \modify{\textbf{Simulation results of COBRA's multi-modal locomotion and associated dynamics.} 
    \textbf{(A) Simulation of locomotion gaits.}  
    \textbf{A1.} \emph{Sidewinding gait} on level terrain: sequential robot configurations with overlaid head-link (purple), tail-link (green) and CoM (blue) trajectories illustrating the characteristic orthogonal body waves.  
    \textbf{A2.} \emph{Vertical undulation (worm) gait}: body-axis sinusoidal waves propagate longitudinally to negotiate narrow passages, with link trajectories shown in green and purple.  
    \textbf{A3.} \emph{Lateral rolling gait}: alternating lateral body contacts generate a rolling progression, with CoM marked by black-white circles and its path in blue.  
    \textbf{A4.} \emph{Hex-ring tumbling locomotion} down a $10^\circ$ incline: the robot reconfigures into a closed ring and performs passive locomotion, with snapshots fading to show successive poses and the CoM descent trajectory (dark blue line).
    \textbf{(B)} Simulated hex-ring tumbling dynamics.  
    Top: $x$–$z$ trajectories of the head link (red), link 6 (blue), and center of mass (magenta) as COBRA tumbles along the incline.  
    Bottom left: normal ground reaction forces for two consecutive links (link 5 in blue, link 4 in red) over a single contact period; shaded backgrounds indicate the active contact intervals.  
    Bottom right: joint torque profiles for four alternating actuators (joint 7 in blue, joint 5 in red, joint 3 in green, joint 1 in purple) over the same time window, highlighting the torque spikes at each tumble event.
    \textbf{(C)} Simulated contact patterns on flat ground.  
    Footprint plots for the sidewinding (left), worm (center), and lateral rolling (right) gaits. Each colored circle denotes the contact location of an individual module, and the solid dark blue curve traces the CoM trajectory.
    \textbf{(D)} Simulated contact forces on flat ground.  
    Vector‐field representations of the normal contact forces during sidewinding (left), worm (center), and lateral rolling (right). Arrows--colored by link index--depict force magnitude and direction at each contact point, with open circles marking the contact positions and an overarching arrow indicating the direction of locomotion.%
}}
\label{fig:sim-plots}
\end{figure}

\modify{
The contact-implicit gait-discovery framework of Sec.~\ref{sec:controls} demands knowledge of link inertias, joint torque limits, ground-contact friction and reliable estimates of state and contact for all twelve modules of COBRA. A high-fidelity MATLAB simulator was therefore developed to reproduce the dominant physics of this contact-rich system before any hardware trials. Each module is modeled as a rigid body of mass m = 0.60 kg with principal moments of inertia I$_\mathrm{x}$ = 7.16707$\times10^{-4}$ kg.m$^2$, I$_\mathrm{y}$ = $8.70397\times10^{-4}$ kg.m$^2$ and I$_\mathrm{z}$ = 8.6286$\times10^{-4}$ kg.m$^2$ derived from the imported STL geometry.

Continuous collision detection operates on the convex hull of each module, which contains approximately 19,000 vertices. Neighboring modules are connected by torque-controlled revolute joints whose motion limits are $\pm$ 70$^\circ$; continuous torque limit is 8.1 Nm and peak torque limit is 10 Nm. A low-level PID controller with gains K$_\mathrm{p}$ = 80, K$_\mathrm{d}$ = 30, K$_\mathrm{i}$ = 10 and an update rate of 1 kHz emulates the embedded low-level controller. Ground reaction is represented by a Hunt–Crossley normal compliance with stiffness k$_\mathrm{n}$ = $10^4$ N/m and damping d$_\mathrm{n}$ = $10^3$ N.s/m, coupled to a smooth stick-slip friction law with coefficients $\mu_\mathrm{s}$=0.7 and $\mu_\mathrm{k}$=0.5 and a critical slip velocity $v_{\mathrm{crit}}$ = $10^{-3}$ m/s. A stiff solver (ODE3) integrated in a fixed-step wrapper of $\Delta t$= 1 ms with absolute and relative tolerances of 1 $\times10^{-3}$ keeps total energy drift low.

Three snake-like gaits synthesized by the trajectory-optimization routine of Sec.~\ref{sec:controls} were evaluated on level ground and are illustrated in Fig.~\ref{fig:sim-plots}A1-3. Sidewinding with a stride period of 2 s achieved an average forward speed of 0.23 $\pm$ 0.1 m/s, a duty factor of 0.325, a peak normal contact force of 219 N and an average mechanical power of 22.6 W. Vertical undulation employed a 2 s gaitcycle and progressed at 0.028 m/s with a duty factor of 0.2085, a maximum normal force of 196 N and a mean power draw of 4.5 W. Lateral rolling required a 2 s period, produced 0.18 m/s, had a duty factor of 0.81, saw peak normals of 279 N and consumed 12.93 W. The corresponding footprint geometries (see Fig.~\ref{fig:sim-plots}C) confirm that sidewinding forms the classic double-helix of contacts, vertical undulation confines contacts to the sagittal mid-line and lateral rolling distributes contacts widely. Contact-force vectors in Fig.~\ref{fig:sim-plots}D remain below the 62 $\mathrm{N\,cm}^{-2}$ bearing-capacity limit of lunar regolith simulants (JSC-1A=1757 kPa, BP-1=1536 kPa, CSM-LHT-1=620 kPa, NU-LHT-4M=1000 kPa, LHS-1=793 kPa).

For crater-wall descent the modules latch into a closed hexagon and are released on a planar 10$^{\circ}$ slope. Figure~\ref{fig:sim-plots}A4 and B show that each roll cycle advances the ring 1.687 m downslope while lowering the CoM by 0.3 m. Seven cycles cover 12 m in 10 s, corresponding to an average speed of 1.2 m/s. Peak normal forces of 80 N on the first-impact link and 104 N on the next occur during 100 ms-long contact intervals. Torque spikes of 8 N\,m at impact remains below 80 \% of the one-second peak rating and continuous torques of 3 N\,m remains below 40 \% of the continuous torque rating.

Further, we explored the effects of tumbling locomotion on granular slopes through high-fidelity simulation using the Discrete Element Method (DEM) \cite{tavarez_discrete_2007,sunday_validation_2019}. Our simulator, based on the DEM Engine from Project Chrono \cite{zhang_2024_deme}, models rigid body-particle interactions by resolving individual contacts between particles and the robot, leveraging GPU acceleration for large-scale computation. DEM applies the rigid-body contact principles described in the path-planning section and captures both inter-particle and particle–object interactions by explicitly resolving large-scale collision forces, friction, and energy dissipation.

We conducted three distinct simulations to compare terrain interaction models:

Simulation 1 - DEM Granular Terrain: The COBRA robot's URDF was allowed to tumble under gravity down a 24° slope. The simulated terrain measured 4 m × 1 m × 0.5 m and was populated with 8000 spherical particles of 5.4 cm diameter, arranged initially in a Hexagonal Close Packing structure. The system was run on a laptop with an Intel Core i5-13420H processor, 16 GB DDR5 RAM, and an NVIDIA RTX 4050M GPU. The terrain was settled for 2 seconds before the robot was released, and the simulation was run with a time step of $5\times10^{-4}$ s. The particle model used friction and restitution coefficients of 0.67 and 0.1 (inter-particle), and 0.67 and 0.2 (particle-robot), respectively, with a rolling resistance of 0.05.

Simulations 2 \& 3 - Simscape Multibody Models: For comparison, we simulated the same tumbling scenario in Simscape Multibody using two different ground models. The first (stiff ground) used a contact model with a spring constant of 1e4 N/m, damping coefficient of 1e3 Ns/m, and a narrow transition width of 1e-3 m to approximate hard ground. The second (compliant ground) used a spring constant of 1e3 N/m, higher damping (1e4 Ns/m), and a broader 0.3 m transition width.

As shown in Fig.~\ref{fig:granular-tumbling}C, both the compliant model and the particle-based simulation resulted in comparable distances traveled, while the stiff ground model allowed the robot to travel significantly farther in the same time. Figure \ref{fig:granular-tumbling}C further plots the normal distance between the robot's CoM and the slope surface (i.e., sinkage). The DEM simulation shows evidence of terrain deformation--particles shift under impact, resulting in dips where the robot partially sinks and bulges where it lifts. In contrast, the stiff ground model produces sharper, more ballistic motion, with brief airborne segments. Normal force plots for a representative period reveal large, short spikes in the stiff model, while the DEM and compliant models spread impact forces over time. These results suggest that DEM captures the dissipative and deformable nature of real granular terrain. Nevertheless, additional experiments and model refinements are required before reliable conclusions can be drawn about mass wasting, sinkage, and traction loss during lunar-crater traversal via tumbling locomotion.

In conclusion, sidewinding on flat terrain thus incurs an energetic cost of roughly 100 J\,kg$^{-1}$\,m$^{-1}$, whereas hex-ring tumbling is nearly passive and therefore preferable for slopes steeper than 8$^\circ$. The strong temporal correlation between simulated torque peaks and the shaded contact states in Fig.~\ref{fig:sim-plots}B shows that it is possible to infer contact onset using only joint-torque residuals and the head-link IMU. These numerical results set safe actuator limits and expected contact signatures for the Lucerne Valley field trials described in Sec.~4.2.}

\begin{figure}
    \centering
    \includegraphics[width=0.8\linewidth]{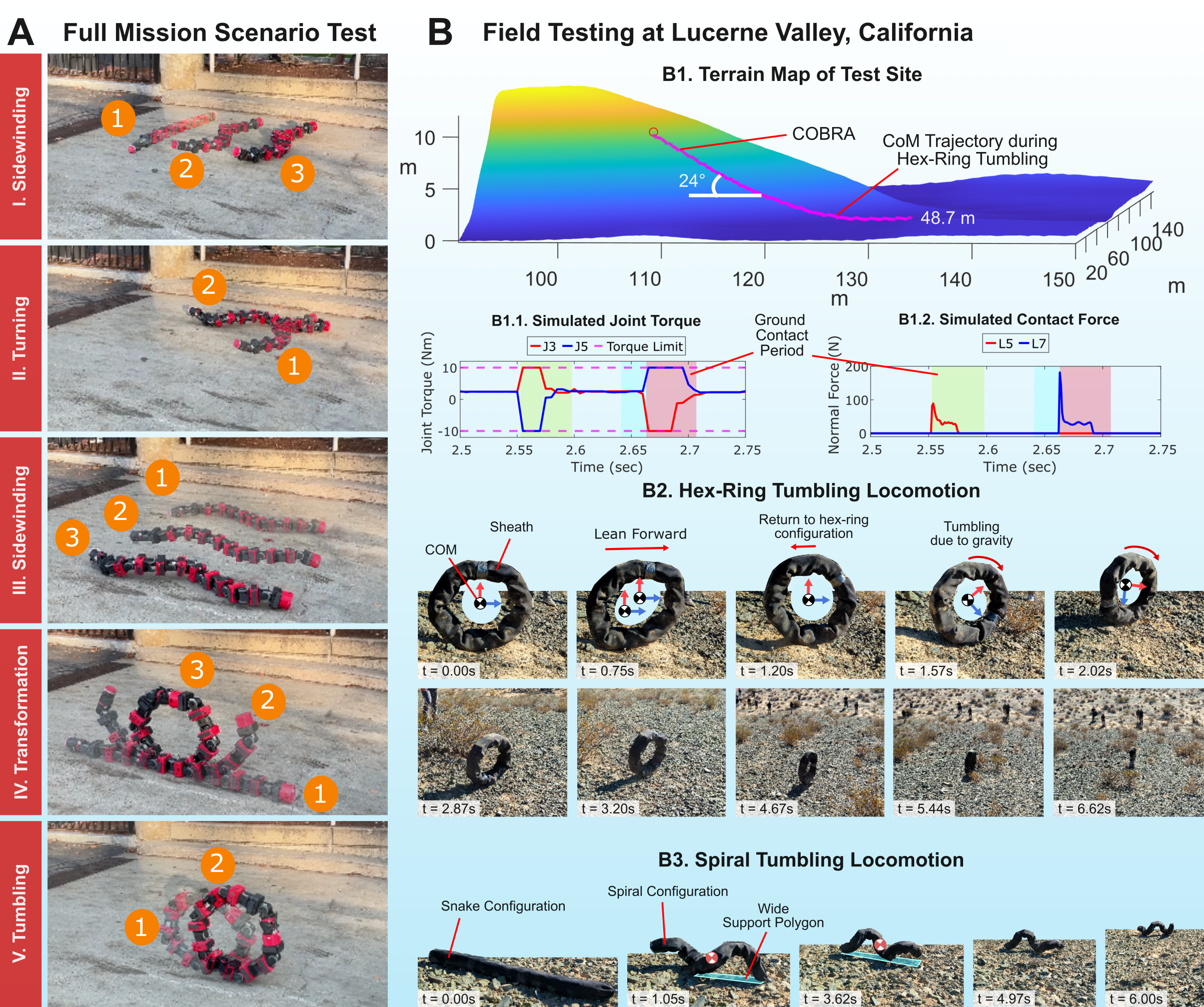}
    \caption{%
    \modify{\textbf{Experimental validation of COBRA's multi‐modal locomotion in indoor and field environments.}
    \textbf{(A)} Full mission scenario test on Northeastern University ramps ($8^\circ$--$10^\circ$ incline): 
    \textbf{(I)} \emph{Sidewinding alignment}: COBRA uses sidewinding to position its longitudinal axis upslope, with labels 1, 2 and 3 marking successive body snapshots. 
    \textbf{(II)} \emph{Turning}: lateral undulation reorients the robot into the rolling direction. 
    \textbf{(III)} \emph{Sidewinding ascent}: the reoriented robot sidewinds straight up the ramp. 
    \textbf{(IV)} \emph{Transformation}: sequential posture manipulation (labels 1–3) closes the hex‐ring configuration. 
    \textbf{(V)} \emph{Tumbling descent}: the closed ring rolls passively down the incline under gravity, then unlatches and returns to snake form to resume sidewinding.
    \textbf{(B)} Field testing at Lucerne Valley, CA demonstrates COBRA's hex‐ring tumbling over a $48.7\mathrm{m}$ natural slope at $24^\circ$ on loose, rocky terrain. Enveloped in a polyester‐fabric sheath, the robot descended in approx. $10~\mathrm{s}$ (avg. $5\mathrm{m/s}$).
    \textbf{(B1)} Terrain map of the test site reconstructed from \cite{usgs_3dep_2021}, with magenta trace showing the COBRA CoM trajectory during tumbling.
    \textbf{(B1.1)} Simulated joint torques for joints 3 (red) and 5 (blue) over consecutive ground‐contact phases; shaded regions denote active contact intervals against the motor torque limit (dashed magenta).
    \textbf{(B1.2)} Simulated normal contact forces at links 5 (red) and 7 (blue), with shaded backgrounds marking their respective contact periods.
    \textbf{(B2)} Field snapshots of hex‐ring tumbling: sequential images at t=0.00, 0.75, 1.20, 1.57, 2.02, 2.87, 3.20, 4.67, 5.44 and 6.62 s, illustrating ring lean, gravity‐driven rollover, and successive tumbles.
    \textbf{(B3)} Spiral tumbling locomotion: the robot transitions from the snake configuration to a helical spiral, forming a wide support polygon (t=0.00, 1.05, 3.62, 4.97 and 6.00 s) to enable stable tumbling on uneven ground.%
}}
    \label{fig:results}
\end{figure}

\subsection{Experiments}

\begin{figure}
    \centering
    \includegraphics[width=0.8\linewidth]{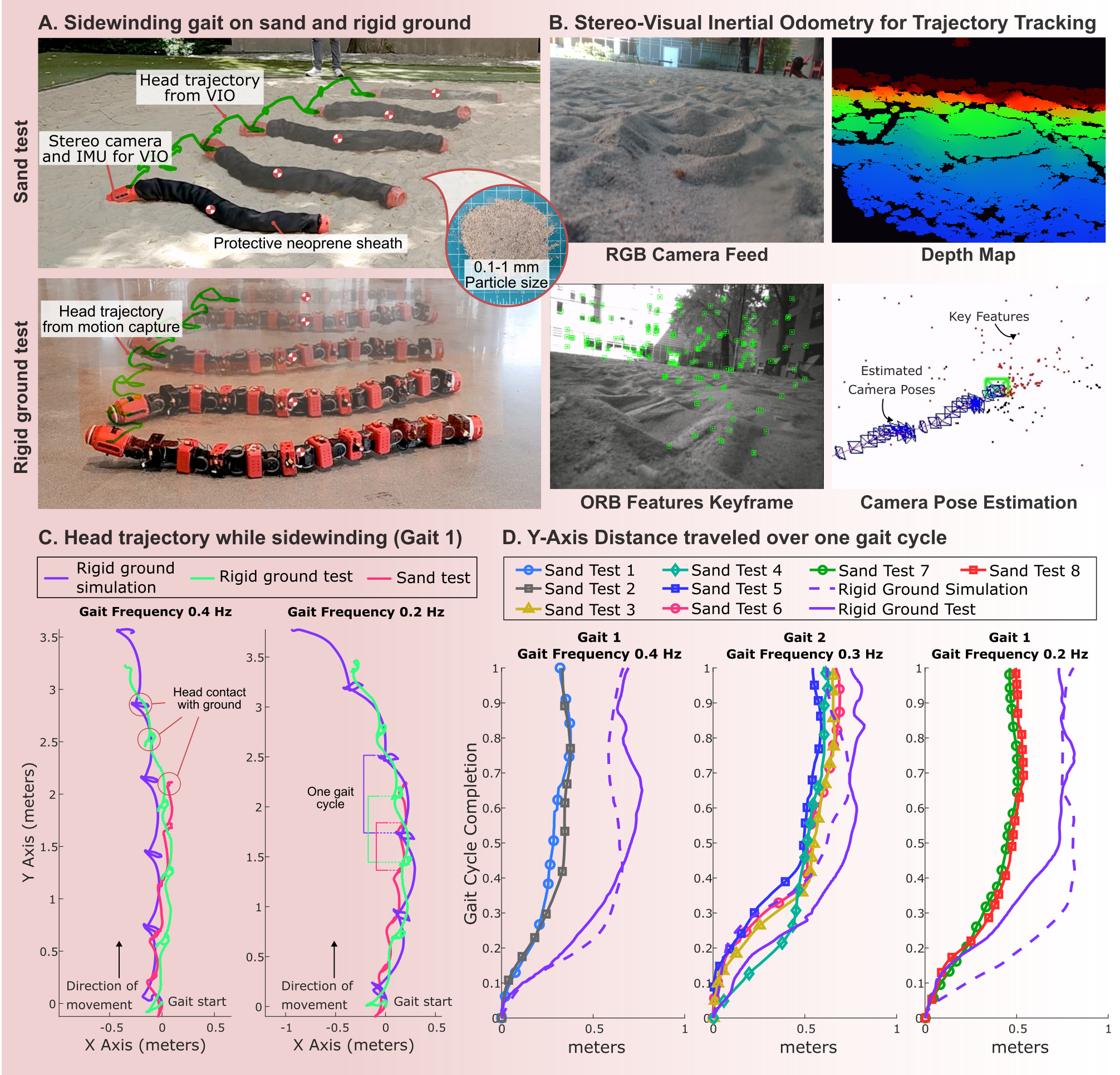}
    \caption{\modify{\textbf{Sidewinding locomotion on sand versus rigid ground and corresponding trajectory analysis.}
    \textbf{(A)} Experimental setup for sidewinding on dry sand (top) at a volleyball court at Northeastern and rigid ground (bottom). The stereo camera-IMU module mounted on the robot's head provides visual-inertial odometry (VIO) on sand for pose estimation, whereas a motion-capture system is used in rigid-ground case. The robot is protected by a neoprene sheath; inset shows the 0.1-1 mm sand grain size.  
    \textbf{(B)} Stereo VIO pipeline used robot pose estimation on sand due to lack of motion capture: RGB image, inferred depth map, ORB key-frame features, and estimated camera poses (coloured point cloud).  
    \textbf{(C)} Planar (X-Y) head trajectories for \emph{Gait 1} at two gait frequencies (0.2-0.4 Hz). Blue indicates a Simscape rigid-ground simulation, green is the rigid-ground experiment, and magenta is the sand test. Head contact points and the progression of one complete gait cycle are highlighted.  
    \textbf{(D)} Normalised Y-axis advance over a single gait cycle \modifyVtwo{for two sidewinding gaits (Gait 1: 40$^\circ$ horizontal, 20$^\circ$ vertical; Gait 2: 60$^\circ$ horizontal, 30$^\circ$ vertical)} at three actuation frequencies (0.2, 0.3, 0.4 Hz). Eight repeated sand trials (distinct colors and markers), the rigid-ground experiment (solid purple), and the rigid-ground simulation (dashed purple) are compared.
    } \modifyVtwo{Only those trials with continuous pose estimates and minimal VIO drift are included after spatio-temporal alignment using the evo package \cite{grupp2017evo}.}
    }
    \label{fig:sand-comparison}
\end{figure}

\modify{
The locomotion modes in Fig.~\ref{fig:locomotion-modes}A were first vetted in simulation and then ported to hardware. In simulation, sidewinding on level ground produced a steady-state velocity of 0.22 m/s at an energetic cost of 100 J\,kg$^{-1}$\,m$^{-1}$, whereas the closed hex-ring descended a \(10^{\circ}\) slope at 5 m/s with peak continuous joint torques of 3 N\,m and peak normal contact forces of 500 N; peak torques are comfortably below the continuous motor rating of 6 N\,m. These margins proved essential when the gaits were migrated to the physical robot, where compliance in the 3-D-printed housings and up to \(10^{\circ}\) of servo backlash produced repeatable misalignment during the head-tail latching phase. Because the angular error was systematic the problem was solved in practice by adding fixed offsets of 3-5$^\circ$ to the final joint commands; the simulation predicted that such a correction would leave torque demand unchanged, a result confirmed empirically. Although a vision-based estimate of the relative head-tail pose would constitute a more general solution, the deterministic offset rendered additional sensors unnecessary for the present study.

With the offset in place the entire locomotion pipeline was exercised on the 8--10\(^{\circ}\) concrete ramps around Northeastern University (Fig.~\ref{fig:results}A). 
At Lucerne Valley COBRA tumbled down a 50 m natural slope at \(24^{\circ}\) (Fig.~\ref{fig:results}B2), reaching the base in \(10\pm1\;\text{s}\) 
for an average speed of 5 m/s. The same terrain profile was imported into the simulator from USGS 3DEP data \cite{usgs_3dep_2021} and the tumbling locomotion was analyzed for peak torque and contact force estimates during tumbling (Fig.~\ref{fig:results}B1.1,\,B1.2). Because the experimental robot carries no foot-plate force sensors, current-derived torques and simulation provide the only practical checks on structural safety.
The Spiral tumbling maneuver of Fig.~\ref{fig:results}B3 further illustrates this point: on the shallow 5\(^{\circ}\) slope the simulated lateral stability margin is retained, predicting tip-over of the hex-ring, exactly what was observed in situ; the helix-shape spiral, whose wider support polygon gives a simulated stability, completed the descent without incident passively compared to the 100 J\,kg$^{-1}$\,m$^{-1}$ mechanical work of sidewinding.

Further, we ran a series of field tests to compare how sidewinding gaits perform on sand versus rigid ground, aiming to understand how different surfaces affect locomotion. The outdoor experiments were conducted on a volleyball court with a natural mix of sand grains—about 20\,\% fine (0.1--0.25 mm), 70\,\% medium (0.25--0.5 mm), and 10\,\% coarse (0.5--1 mm). To protect the robot from dust during outdoor trials, we used the neoprene sheath and sealed the head module, leaving only the stereo camera exposed. We tracked the robot's head trajectory using ORB-SLAM3 \cite{campos_orb-slam3_2021} 
with stereo-inertial data. For comparison, we used motion capture in the lab to track the same gaits on rigid ground. Figure~\ref{fig:sand-comparison}A shows snapshots from both environments, while Fig.~\ref{fig:sand-comparison}B illustrates the view from the onboard camera along with SLAM feature tracking and estimated poses.

We compared two sidewinding gaits with different contact patterns at three different frequencies on both surfaces. Figure~\ref{fig:sand-comparison}C shows the tracked trajectories for gait 1. As expected, the simulation traveled the farthest, since it does not include actuator errors and assumes ideal contact. In the hardware tests, the robot covered more ground on the rigid surface than on sand, where higher slippage reduced overall effectiveness. Still, the general gait pattern remained consistent across both terrains. This becomes clearer in Fig.~\ref{fig:sand-comparison}D, which normalizes the trajectories over one gait cycle. On sand, each cycle resulted in shorter displacement, but the gap was smaller in gait 2, where the contact pattern favors more stable, quasi-static support. Slowing the gait down also reduced slippage on both surfaces, leading to more distance per cycle--though, naturally, it took more time to complete.

To conclude, taken together, the numerical and field data show a consistent hierarchy of modes: sidewinding is reliable but slow and energy-intensive on gentle grades, spiral tumbling extends the range to shallow inclines while keeping actuator loads low, and hex-ring tumbling is the fastest and least energy-dense option once the slope exceeds approximately \(8^{\circ}\). The simulator not only anticipated these cross-overs but also provided quantitative limits on torque, contact force and alignment error that proved robust when transferred to hardware, thereby fulfilling its role as an indispensable precursor to lunar deployment.
}

\section{Concluding Remarks}

Rover exploration of the lunar surface has been a standardized approach by NASA. However, these systems have not been previously attempted in the challenging environment of the lunar craters due to the significant risk of immobilization. With the Artemis program's mission to revitalize lunar and space exploration, the need for efficient and fast locomotion concepts has never been greater. Particularly, since one Artemis objective is to create a sustainable human basecamp on the Moon, ISRU on the Moon such as the search for water ice, which can potentially supply drinking water, oxygen, and rocket propellant, becomes immensely important. Currently, the means to access the ice deposits on the Moon are minimal. 

\modify{
In this report, we summarized our efforts to design, prototype, and evaluate a multi-modal snake-style robot called COBRA, Crater Observing Bio-inspired Rolling Articulator. Specifically, the technical contributions of this work are: (1) Hardware design, including head, tail, and body module mechanical and electronics design. (2) Design of the head-tail docking mechanism to substantiate transitions between crawling and tumbling. (3) Numerical modeling of crawling and tumbling locomotion feats based on Lagrange and mixed Hamilton-Lagrange (partitioned state-space model for tumbling) dynamics. (4) Contact-implicit optimization-based gait discovery and joint motion planning. (5) Simulation and experimental validation of hex-ring tumbling, spiral tumbling, sidewinding, lateral rolling, and vertical undulation. (6) Demonstration of field-tested locomotion modes in dusty, steep, and bumpy environments, with self-sustained dust mitigation. (7) Demonstration of full deployment from a lander, including unmanned transitions between various locomotion modes in experiments.

We demonstrated COBRA's mobility beyond conceptual functions and displayed promising mobility solutions for environments analogous to Shackleton Crater. With the demonstrated capabilities, our estimation of COBRA's Technology Readiness Level (TRL) is 5-6 \cite{tzinis_technology_2015}. Our path-to-flight evaluation, partly covered in this paper and with additional capabilities to be shown in subsequent papers, demonstrates that the COBRA concept has moved far beyond conceptual functions and now offers promising mobility solutions in environments analogous to Shackleton Crater. COBRA can now traverse flat surfaces using different gaits (sidewinding, vertical undulation, lateral rolling, etc.), move over sloped surfaces, operate in dusty environments, and carry onboard sensors and computers for autonomous operations. Other capabilities yet to be reported in future papers include grasping objects, object manipulation, and locomotion on both flat and steep slopes.

The fully integrated COBRA prototype has demonstrated essential navigation capabilities both inside and outside controlled laboratory settings, including traversal on angled concrete and sand-covered terrain. However, while laboratory environments have been the focus, future work should prioritize outdoor experiments. That said, field-tested Hex-Ring tumbling, Spiral tumbling, and sidewinding capabilities in dusty, contested, and unstructured environments have been reported as part of this work. To advance beyond this stage, the next steps involve first exploring space-proven sheath materials, such as Vectran and Kevlar, and conducting performance testing under conditions simulating the space environment (e.g., thermal vacuum and bakeout tests) to validate compatibility with CLPS landers.

Second, for individual subsystems, including the joints and modules, larger payload capacity should be considered so that space-worthy metal-printed modules replace the current Markforged-printed components. Currently, the joint attachments--serving as the chassis for add-on geometries and ensuring secure connections between modules--are designed for smaller payloads, such as carrying batteries and power electronics. Adding additional payloads beyond this design limit could negatively affect the robot's performance. For future developments, a single joint module encapsulating the voltage regulator, motor, battery, and a metal housing should undergo bench testing to simulate radial and axial loads, ensuring the integrity of its structural design. To prevent axial loads from exceeding actuator specifications, stronger high-duty needle thrust bearings should be integrated.

Third, from a locomotion perspective, the successful implementation and consistency of the latch mechanism under different operationally contested scenarios for hexagonal tumbling demonstrate progress consistent with TRL 6. This mechanism enabled COBRA to perform tumbling on a 24-degree slope, initiated by shifting its center of mass. However, future testing will involve steeper and more granular slopes, as well as continuous closed-loop feedback-driven control, to validate the mechanism under more challenging conditions. Steering the heading angle during tumbling through posture manipulation remains an ongoing research problem, and research results on this front will be reported in subsequent publications.

Lastly, COBRA requires the integration of a comprehensive sensor package to enable data acquisition from the payload module during field operations. Expanding the sensor suite, developing robust path-planning algorithms, and implementing object detection software are crucial next steps for achieving end-to-end software functionality. Additionally, incorporating sophisticated exception-handling logic and autonomous verification of motor performance will ensure predictable operational capabilities. With these evaluations and a clear roadmap, we envision deploying COBRA for future explorations of Shackleton Crater, targeting safe and reliable navigation and data acquisition in complex lunar environments. With these improvements, COBRA will be on a clear trajectory toward deployment for Shackleton Crater exploration, enabling scientific data collection in complex lunar terrains.}

\medskip
\textbf{Supporting Information} \par 
Supporting Information is available from the Wiley Online Library or from the author.

\medskip
\textbf{Acknowledgements} \par 
This work was carried out with support from NASA Space Technology Mission Directorate’s Game Changing Development Program as part of the 2022 NASA BIG Idea Competition \modify{ and a National Science Foundation (NSF) CAREER Award (Award No. 2340278). This work was also supported by a NASA Space Technology Graduate Research Opportunity (Grant No. 80NSSC24K1393).}

We acknowledge Neha Bhattachan, Yash Bhora, Andre Caetano, Ian McCarthy, Brandon Petersen, Alexander Qiu, Matthew Schroeter, Nolan Smithwick, Konrad Sroka, Madeleine Weaver, Jason Widjaja for their key contributions to the design and development of COBRA.

\medskip
\textbf{Contributions} \par
A.S. did simulation, experiments, figure preparation; H.N. assisted with the preparation of Fig.~\ref{fig:cobra} and Fig.~\ref{fig:locomotion-modes}; A.S., A.R. collaboratively wrote the paper and edited the draft. H.N., E.S., A.K. reviewed the paper and provided suggestions. The mathematical framework is by A.R.

\medskip

%
\printbibliography
\newpage
\begin{figure}
  \includegraphics{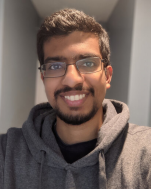}
  \caption*{Biography: Adarsh Salagame is a PhD candidate in Computer Engineering at Northeastern University, focusing on perception and control for bio-inspired multimodal robots. His research includes real-time state estimation, motion planning, and optimal control for autonomous systems. He holds an MS in Robotics from Northeastern and has research experience building and controlling aerial, underwater and multimodal robots.}
\end{figure}

\begin{figure}
  \includegraphics{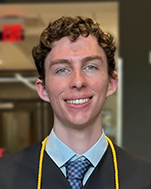}
  \caption*{Biography: Henry Noyes is a graduate student at Northeastern University pursuing a Master of Science in Robotics. His research focuses on advancing autonomy in bio-inspired snake robots for lunar exploration. He graduated with a Bachelor of Science in Mathematics and Physics, with a minor in Data Science, from Northeastern. His interests include space robotics, autonomous systems, and bio-inspired locomotion for extraterrestrial environments.}
\end{figure}

\begin{figure}
  \includegraphics{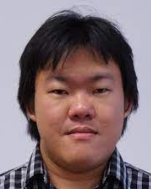}
  \caption*{Biography: Dr. Eric Sihite is a Postdoctoral Research Associate at Caltech. He received his Ph.D. in Mechanical Engineering from the University of California San Diego, where he focused on robotics and control systems. His research includes the mechanical design, system dynamics, and control of miniaturized ball-balancing robots and flapping-wing aerial drones. He has expertise in system modeling, state estimation, controller design, and numerical simulations.}
\end{figure}

\begin{figure}
  \includegraphics{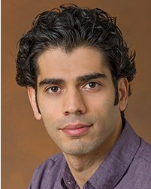}
  \caption*{Biography: Dr. Arash Kalantari is a Robotics Mechanical Engineer at JPL. He specializes in mechanisms and mechanical design of robotic systems. He graduated with a Ph.D. in Mechanical and Aerospace Engineering from Illinois Institute of Technology in 2015. The focus of his PhD research was on "Multi-Modal Mobile Robots.” The HyTAQ, Hybrid Terrestrial and Aerial Robot, and ResQuake are examples of the robotic platforms he has developed.}
\end{figure}

\begin{figure}
  \includegraphics{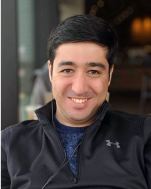}
  \caption*{Biography: Dr. Alireza Ramezani is an Associate Professor of Electrical and Computer Engineering at Northeastern University. He earned his PhD in Mechanical Engineering from the University of Michigan in 2014, following an MS from ETH Zurich in 2010 and a BSc from the Iran University of Science and Technology in 2007. Dr. Ramezani's research focuses on designing bio-inspired robots with complex morphologies, emphasizing nonlinear control systems and diverse locomotion strategies.}
\end{figure}



\end{document}